\documentclass[pdflatex,sn-mathphys-ay]{sn-jnl}

\usepackage{longtable}
\usepackage{graphicx}%
\usepackage{multirow}%
\usepackage{amsmath,amssymb,amsfonts}%
\usepackage{amsthm}%
\usepackage{mathrsfs}%
\usepackage[title]{appendix}%
\usepackage{xcolor}%
\usepackage{textcomp}%
\usepackage{manyfoot}%
\usepackage{booktabs}%
\usepackage{algorithm}%
\usepackage{algorithmicx}%
\usepackage{algpseudocode}%
\usepackage{listings}%
\usepackage{latexsym}
\usepackage{array} 
\usepackage{color}
\usepackage{mathtools}
\usepackage{microtype}
\usepackage{subfigure}

\newcommand{\BibTeX}{B\kern-.05em{\sc i\kern-.025em b}\kern-.08em\TeX}

\DeclareMathOperator*{\argmax}{arg\,max}

\theoremstyle{thmstyleone}%
\theoremstyle{thmstyletwo}%

\theoremstyle{thmstylethree}%

\begin{document}

\title[Problem-oriented AutoML in Clustering]{Problem-oriented AutoML in Clustering}


\author*[1,2]{\fnm{Matheus} \sur{Camilo da Silva}}\email{matheus.camilodasilva@phd.units.it}

\author[3,4]{\fnm{Gabriel} \sur{Marques Tavares}}\email{tavares@dbs.ifi.lmu.de}

\author[2]{\fnm{Eric} \sur{Medvet}}\email{emedvet@units.it}

\author*[2]{\fnm{Sylvio} \sur{Barbon Junior}}\email{sylvio.barbonjunior@units.it}

\affil*[1]{\orgdiv{Applied Data Science \& Artificial Intelligence}, \orgname{University of Trieste}, \orgaddress{\street{Piazzale Europa, 1}, \city{Trieste}, \postcode{34127}, \state{Friuli-Venezia Giulia}, \country{Italy}}}

\affil[2]{\orgdiv{Department of Engineering and Architecture}, \orgname{University of Trieste}, \orgaddress{\street{Piazzale Europa, 1}, \city{Trieste}, \postcode{34127}, \state{Friuli-Venezia Giulia}, \country{Italy}}}

\affil[3]{\orgdiv{Database Systems and Data Mining}, \orgname{LMU Munich}, \orgaddress{\street{Oettingenstraße 67}, \city{Munich}, \postcode{80538}, \state{Bavaria}, \country{Germany}}}

\affil[4]{\orgname{Munich Center for Machine Learning (MCML)}, \orgaddress{\city{Munich}, \state{Bavaria}, \country{Germany}}}


\abstract{The Problem-oriented AutoML in Clustering (PoAC) framework introduces a novel, flexible approach to automating clustering tasks by addressing the shortcomings of traditional AutoML solutions. Conventional methods often rely on predefined internal Clustering Validity Indexes (CVIs) and static meta-features, limiting their adaptability and effectiveness across diverse clustering tasks. In contrast, PoAC establishes a dynamic connection between the clustering problem, CVIs, and meta-features, allowing users to customize these components based on the specific context and goals of their task. At its core, PoAC employs a surrogate model trained on a large meta-knowledge base of previous clustering datasets and solutions, enabling it to infer the quality of new clustering pipelines and synthesize optimal solutions for unseen datasets. Unlike many AutoML frameworks that are constrained by fixed evaluation metrics and algorithm sets, PoAC is algorithm-agnostic, adapting seamlessly to different clustering problems without requiring additional data or retraining. Experimental results demonstrate that PoAC not only outperforms state-of-the-art frameworks on a variety of datasets but also excels in specific tasks such as data visualization, and highlight its ability to dynamically adjust pipeline configurations based on dataset complexity.}


\maketitle

\section{Introduction}\label{sec1}
Most Machine Learning (ML) applications are composed by a sequence of steps (i.e., a pipeline) involving the selection of algorithms and the setting of their hyperparameters. However, carefully designing good performing pipelines is a tedious, error-prone task that often relies on expert knowledge, which is frequently unavailable or expensive\citep{Olson2016AutomatingBD}. Automated Machine Learning (AutoML) provides the tools to make the creation and configuration of optimal ML solutions more accessible to practitioners with varying levels of expertise by automating the complex and time-consuming aspects of ML development \citep{hutter2019automated}. 

This automation is particularly straightforward for supervised tasks where labeled data is available, allowing for objective evaluation metrics like accuracy, precision, or F1 score to guide optimization. By systematically testing different pipeline configurations, AutoML can identify the most effective solutions for a given dataset \citep{he2021automl, truong2019towards}. 
 
In the case of unsupervised tasks, the lack of labels, pose a challenge for the evaluation of generated pipelines. A conventional approach is the reliance on internal Clustering Validity Indexes (CVI) to synthesize and optimize clustering solutions \citep{bahri2022automl}, while ignoring any external information of the clustering problem. This approach's only concern is to optimize CVIs \citep{Kryszczuk2010EstimationOT, Marutho2018TheDO, tschechlov2021automl4clust}. This practice however can be limiting as each internal score evaluates only a facet of the relationships between data points in a clustering setting. Moreover, different evaluation metrics may lead to different conclusions about the quality of a clustering solution \citep{rokach2005clustering}. For this reason, properly defining the clustering problem (i.e., the goal of the clustering analysis) and a corresponding optimization function is crucial for generating high quality automated clustering solutions \citep{luxburg2012clustering}.
 
Another approach in AutoML for clustering is to create solutions based on Meta-learning, focusing on leveraging prior experience to enable models to quickly adapt to new tasks by learning from past clustering problems and using this knowledge to inform future decisions \citep{brazdil2022meta, hutter2019automated}. By accumulating insights from diverse datasets, meta-learning provides a way to transfer knowledge, thus reducing the time and computational resources needed to optimize new tasks. However existing works, such as \citep{poulakis2020autoclust, liu2021autocluster, elshawi2021csmartml, treder2023ml2dac}, tend to restrict their application of meta-learning by attaching a fix set of algorithms, hyperparameter configurations, and CVIs to their meta-database, which can hinder generalization. This presents an opportunity for more flexible solutions that can adapt across a wider variety of clustering tasks and optimization objectives.

Since there is not a universal metric capable of describing every clustering goal equally well \citep{bonner1964some}, the definition of clustering quality can be subjective and dependent on the context of the problem that the practitioners conducting the clustering aim to solve \citep{bonner1964some}. That is, the clear association between the clustering task at hand and suitable metrics to measure performance should be considered when developing clustering solutions \citep{luxburg2012clustering,hennig2015what,mechelen2023white}. Therefore, a partitioning that is optimal for data visualization could be inadequate for noise reduction.

The Problem-oriented AutoML in Clustering (PoAC) framework introduces a methodology that directly addresses the limitations of traditional AutoML solutions by establishing a dynamic connection between the clustering problem, CVIs, and meta-features. PoAC allows flexibility in adapting these components to the specific needs of the clustering task at hand. This flexibility enables users to customize both the meta-features and CVIs based on their problem's context, ensuring that the solution aligns with their particular objectives. At the core of PoAC is a surrogate model, trained on a large meta-knowledge base of previous clustering datasets and their associated solutions. This surrogate model is then leveraged to infer the quality of new clustering pipelines, making it possible to synthesize optimal solutions for unseen datasets while accounting for the user’s clustering goals.

This flexibility is a key differentiator of PoAC compared to traditional AutoML frameworks, which often offer generalized solutions with fixed CVIs and meta-features. While such general-purpose approaches aim to be widely applicable, they can become overly restrictive, providing solutions that only make sense in the context of the predefined CVIs they include. In contrast, PoAC’s ability to adjust CVIs and meta-features ensures that the clustering pipeline is specifically tailored to the user's intent, thus delivering a more problem-oriented solution. By bridging the gap between the characteristics of the dataset and the quality metrics used to evaluate clustering, PoAC offers a more nuanced, data-driven approach to AutoML that reflects the practical goals of clustering practitioners. PoAC is algorithm-agnostic, generating solutions based on the available set of algorithms without demanding additional training or data ingestion. It is entirely decoupled from the optimization approach. 

Results showed that PoAC obtained superior performance when compared to state-of-the-art frameworks on a set of diverse datasets, while also obtaining better results on CVI's related to the visualization problem. Furthermore, experiments demonstrate PoAC's capacity to dynamically synthesize pipelines, by adding or removing preprocessing steps depending on the complexity of the dataset and the defined problem.

\section{Theoretical Background}\label{sec:theoretical_background}

AutoML refers to the application of techniques to automate the process of designing and optimizing ML pipelines \citep{hutter2019automated}, i.e., acting in the intersection of automation and ML \citep{yao2019taking}. Considering the plethora of algorithms and techniques for each step of a ML pipeline, the configuration space grows exponentially \citep{hutter2019automated}. In practical scenarios, the abundance of options available for pipeline development introduces significant complexity, posing challenges even for seasoned experts. The conventional approach to this complexity involves iterative processes, frequently relying on trial-and-error methodologies \citep{zoller2021benchmark}. Unfortunately, this process is inherently resource-intensive and time-consuming, often resulting in sub optimal pipelines due to the limited exploration of potential configurations. 

AutoML literature has developed specific methods to solve different optimization problems, that is, breaking down the pipeline search to smaller, simpler tasks. The most elementary AutoML task is Algorithm Selection (AS). Given a portfolio of algorithms and a dataset, AS is the task of selecting the algorithm that maximizes performance for the input data. AS was first proposed and discussed by \citep{rice1976algorithm}, where he formalizes the AS problem as the mapping of a \textit{problem space} (represented by datasets for a specific problem) to a performance measure space that evaluates pipeline efficiency. This association might come in the form of meta-features describing the \textit{problem space} behavior, therefore, mapping the \textit{problem space} to a feature space, which, in turn, is correlated with an algorithm space containing potential solutions. The quality of algorithms is measured by the performance space. The ultimate goal is to find a mapping function that optimally associates problems with performances (going through algorithms), thereby maximizing performance for a given task.

Furthermore, algorithms are configured by their hyperparameters, which significantly affects performance in clustering settings \citep{mishra2022evaluative}. In other words, optimizing the hyperparameters leads to improved generalizations and, consequently, better performances. This task is coined as Hyperparameter Optimization (HPO) \citep{hutter2019automated}. HPO is conceptually similar to AS, however, hyperparameters often lie in continuous and high dimensional spaces, making the \textit{search space} considerably larger. Hence, HPO is often more challenging than AS. While AS focuses on finding an optimal algorithm and HPO on selecting optimal hyperparameter configurations, both tasks can be joined. The Combined Algorithm Selection and Hyperparameter Optimization (CASH) task comprises both previously presented problems \citep{hutter2019automated}. Although the \textit{search space} is expanded even more, CASH solutions tend to outperform others as they combine crucial aspects of pipeline design that directly affect performance. As a natural extension, we also introduce here the Pipeline Synthesis (PS) task. Since CASH is only focused on algorithms and hyperparameters, PS extends the concept to additional pipeline steps, such as preprocessing, data transformation, feature selection, dimensionality reduction, among others. Therefore, in the scope of this research, PS refers to the search of optimal pipelines for a given clustering problem.

\section{Related Works}
\label{RW}

The exploration of AutoML applications specifically tailored for clustering problems has gained some prominence \citep{ferrari2015clustering,pimentel2019new,souto2008ranking,pimentel2018statistical,nascimento2009mining,ferrari2012clustering,soares2009analysis,fernandes2021towards,pimentel2019unsupervised,gabbay2021isolation}. For that, many apply meta-learning techniques, which are known to perform quite well for AS. However, they are only able to select an algorithm out of a portfolio. Considering (i) the complexity of real environments (i.e., necessity of configuring several pipeline steps), (ii) the importance of tuning hyperparameters, specially for clustering, and (iii) that our solution searches for the complete pipeline, we limit the scope of our related work to research that tackles at least the CASH problem.

\citep{poulakis2020autoclust} presented a framework for clustering problems based on a meta-learning approach. For this, the authors first build a meta-database containing information about datasets behavior (captured by meta-features) combined with CVIs measured after the application of several clustering algorithms, which are known as landmarking meta-features. The meta-learning step recommends the algorithm to be used for a new dataset and is coupled with Bayesian optimization to tune the hyperparameters. Similarly, \citep{liu2021autocluster} also builds a meta-database comprising clustering solutions and performance measurements. Moreover, the authors consider a multi-objective function as the optimization goal, which makes the problem more complex. Once the algorithm is selected, the framework applies grid search as a strategy for HPO. It has been pointed out, however, that grid search is the most naive optimization approach, often leading to non-optimal solutions \citep{hutter2019automated}. \citep{elshawi2021csmartml} proposed a solution that also uses meta-learning, but instead of only recommending the algorithm, it selects the appropriate CVI to be optimized considering dataset characteristics. Once algorithm and optimization goals are set, the method uses evolutionary algorithms for HPO. However, selecting different CVIs for each dataset leads to solutions that are not comparable as the goal is not unique. Finally, \citep{cohenshapira2021automatic} used meta-learning based on supervised graph embedding optimized towards clustering AS. The selected algorithm has its hyperparameters optimized using Bayesian optimization.

A common characteristic of previous solutions is the use of meta-learning for AS, which is in accordance with state-of-the-art in supervised AutoML. However, a clear limitation of such approaches is breaking down the CASH problem into first solving AS and, finally, HPO. Dividing CASH into two steps often leads to local optima, therefore, solutions might not achieve a suitable performance. This phenomenon is observed because once the search is narrowed down into only one algorithm, all the other candidate solutions in the portfolio are dropped down prematurely in the search. As clustering algorithms are heavily affected by hyperparameter configuration, algorithms that were ignored too early in the search might have achieved optimal performance after HPO.

A few solutions tackled the CASH problem directly (i.e., optimizing both algorithms and respective hyperparameters together) \citep{craenendonck2017constraint,tschechlov2021automl4clust,treder2023ml2dac}. \citep{craenendonck2017constraint} proposed a constraint-based approach where users input their restrictions and used grid search as the optimization mechanism. \citep{tschechlov2021automl4clust} introduced a framework with several optimization techniques such as random search, Bayesian optimization and Hyperband. However, it is left to the user to choose the preferred technique along with the CVI to be optimized, going in the opposite direction of AutoML, where the aim is to release the burden of design choices from users. Finally, \citep{treder2023ml2dac} also presented a framework based on meta-learning for warm-starting and CVI recommendation, which is then coupled with Bayesian optimization methods for HPO. However, as the authors showed, clustering performance for external datasets is poor. To the best of our knowledge, the only approach tackling the PS task was proposed by \citep{elshawi2022tpe}. For that, the authors create a three stepped framework: meta-learning for warm-starting, evolutionary algorithms for HPO and clustering ensemble to provide a solution. However, the \textit{search space} is quite limited, including only a few preprocessing techniques.

Although there are interesting approaches in the literature, specially from the engineering point of view, all of them ignore the theoretical limitations of proposing generic clustering solutions without considering the clustering goal. As pointed by \citep{luxburg2012clustering}, one must consider the objective (i.e., for which reason the dataset is being clustered) in order to build a clustering solution. The same applies for AutoML-based clustering frameworks, providing a fixed set of CVIs to be optimized for any given clustering task completely ignores the subjectivity inherently inserted in clustering analysis. Considering previous limitations, our approach is the first one that proposes a problem-oriented AutoML framework to synthesize clustering pipelines.

\section{Problem-oriented AutoML in Clustering (PoAC)}
\label{Methods}

\begin{figure}[!htp]
    \centering
    \includegraphics[width=1\textwidth]{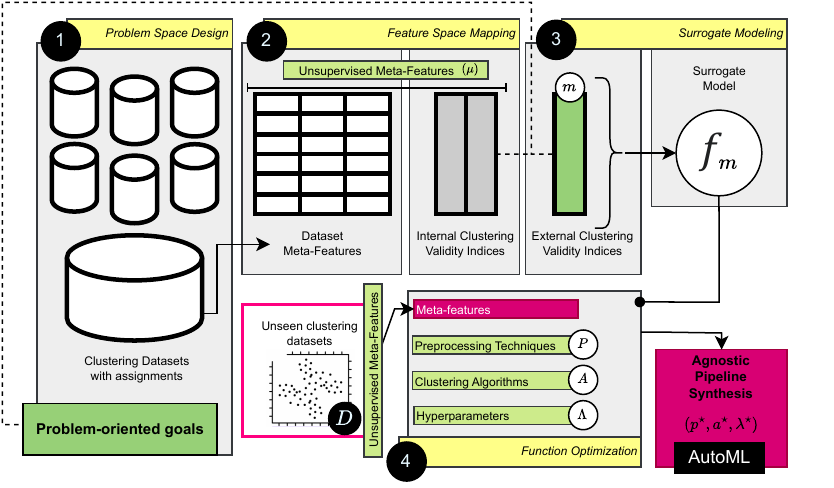}
    \caption{Problem-oriented AutoML in Clustering (PoAC) framework composed by \textit{problem space} design, \textit{feature space} mapping, surrogate modeling, function optimization and pipeline recommendation.}
    \label{fig:method}
\end{figure}

An overview of the proposed approach is displayed in \autoref{fig:method}. It consists of four main stages: Problem Space Design, Feature Space Mapping, Surrogate Modeling, and Function Optimization. Each of these stages is discussed in detail below.

The \textbf{first stage} is the composition of a problem space with labeled clustering datasets related to specific goals, such as visualization, denoising, dimensionality reduction, etc. This foundational process aims to assemble a sizable and diverse problem space that encapsulates a broad spectrum of clustering challenges, patterns, and complexities. 

In the \textbf{second stage}, the datasets are mapped to a feature space, allowing PoAC to build an enriched knowledge representation for surrogate model induction. The enriched feature vector is composed of unsupervised meta-features extracted from the datasets and CVIs. These CVIs are derived by adding increasing levels of noise to the cluster labels of each dataset, providing a range of performance values.

The meta-features provide a higher-level representation of characteristics extracted from the datasets, capturing essential information about their structure, complexity, and statistical properties, as well as from pipeline candidate solutions.

The \textbf{third stage} is the training of a surrogate model to predict a continuous target value (external CVI). In this scenario, it is possible to employ various regression algorithms (or ensembles) for training the surrogate model. By leveraging regression algorithms for the surrogate model, one can empower it with the capacity to discern intricate relationships between meta-features, internal and external CVIs, paving the way for a robust and task-specific automated PS in the final step of the PoAC framework.

The \textbf{fourth stage} is the stage where a given optimization method takes advantage of the trained surrogate model as a objective function to synthesize complete pipelines for clustering according to the specific problem defined in the first stage.

\subsection{Problem statement: surrogate-based PS for clustering}
We define a \emph{dataset} $D=\{x^{(i)}\}_{i=1}^{i=n} \in \mathcal{P}(X)$ and $X$ as a set of observations, each observation being $x^{(i)} \in X$, and $n$ is the cardinality of $X$.

We define a \emph{problem} $u \in U$ as a user-defined clustering goal to partition the data $D$.

We define \emph{meta-features} $\mu: \mathcal{P}(\mathcal{P}(X)) \to \mathbb{R}^{l}$ as a function that maps a dataset $D \in \mathcal{P}(X)$ to a $l$-dimensional numerical space.

We define a \emph{preprocessing technique} $p: \mathcal{P}(X) \to \mathcal{P}(X')$ as a function that maps a dataset $D \in \mathcal{P}(X)$ to another dataset $p(D, \Lambda_p) \in \mathcal{P}(X')$; we denote by $P_{X,X'}$ the set of preprocessing techniques compatible with $X$ and $X'$; we denote $\Lambda_p$ as the set of hyperparameters associated with the selected preprocessing method.

We define a clustering \emph{algorithm} as function $a: \mathcal{P}(X') \to \mathcal{P}(\mathcal{P}(X'))$ that takes a dataset $D \in \mathcal{P}(X')$ and returns a partition $a(D, \Lambda_a) \in \mathcal{P}(\mathcal{P}(X'))$ of $D$, i.e., a set of subsets of $D$ such that $\bigcup_{D' \in a(D)} D' = D$ and $\forall D',D'' \in a(D), D' \cap D'' = \emptyset$; we denote by $\Lambda_a$ the set hyperparameters associated with the selected algorithm;

We define a \emph{CVI} $m: \mathcal{P}(\mathcal{P}(X')) \to \mathbb{R}$ as a function that measures the quality of a partition---for simplicity, we assume that for the values of $m$, the greater, the better; we denote by $M_X$ the set of CVIs compatible with $X$.

We define a clustering \emph{pipeline} over $X$ a tuple $((p_1, \Lambda_{p_{1}}), \dots, (p_k, \Lambda_{p_{k}}), (a, \Lambda_{a}))$ such that $(p_1, \Lambda_{p_{1}}) \in P_{X,X_1}$, $(p_2, \Lambda_{p_{2}}) \in P_{X_1,X_2}$, \dots, $(p_k, \Lambda_{p_{k}}) \in P_{X_{k-1},X_k}$ is a sequence of mutually compatible preprocessing techniques and their corresponding hyperparameters and $(a_1, \Lambda_{a_{1}}) \in A_{X_k}$ is a clustering algorithm and its corresponding hyperparameters compatible with the output of $(p_k, \Lambda_{p_k})$, i.e., $X_k$.
For the sake of clarity and without a practical loss of generality, provided that the used $X$ is general enough, we consider, from now on, only clustering pipelines where all the elements (preprocessing techniques and clustering algorithm) operate on the same $X$.
Under this assumption, and omitting the subscript $X$, the set of all clustering pipelines defined over $X$ is $P^* \times A \times \Lambda_{P^*} \times \Lambda_{A}$, where $P^*$ denote the sequence of elements of $P$.

Given a CVI $m \in M$ and a dataset $D \in \mathcal{P}(X)$, finding the best pipeline $(p^\star,a^\star,\lambda^\star)$ corresponds to solving the following optimization problem:
\begin{align}
    (p^\star,a^\star,\lambda^\star)
    &= \argmax_{(p,a,\lambda) \in P^* \times A \times \Lambda} (m \circ a \circ p)(D,\lambda_a,\lambda_p) \nonumber
\end{align}
where $p = \{p_1,\dots,p_k\}$ is a sequence of preprocessing techniques, which can be empty, and the operator $\circ$ denotes the composition of functions.

It can be seen that solving the optimization problem defined above by enumeration would require actually applying each candidate pipeline to the dataset $D$.
We here propose to use a \emph{surrogate function} $f_m: P^* \times A \times \Lambda \times \mu(\mathcal{P}(X)) \to \mathbb{R}$ that, given a pipeline $(p,a,\lambda)$ and a dataset $D$, approximates the value of the CVI $m$ computed on the partition obtained applying the pipeline on the dataset, i.e.:
\begin{equation}
    f_m((p,a, \lambda), \mu(D)) \approxeq (m \circ a \circ p)(D)
\end{equation}
Solving the problem of finding the best pipeline for a CVI $m \in M$ and a dataset $D \in \mathcal{P}(X)$ can be hence cast to optimizing $f_m$:

\begin{equation}
    \argmax_{(p,a,\lambda) \in P^* \times A \times \Lambda} f_m((p,a,\lambda), \mu(D), u)
\end{equation}

\section{PoAC for the visualization problem}\label{sec:setup}

In this work, we consider the visualization as the goal of a PoAC instance. Recently, the relevance of visualization for the interpretation of ML methods is increasing \citep{chatzimparmpas2020survey}, specially in unsupervised learning, where it provides a unique lens through which complex data structures can be interpreted. The objective of clustering for visualization is to uncover meaningful patterns or groupings within datasets and represent them in a visually comprehensible manner \citep{ezugwu2022comprehensive,al2019computational}. The aim is not only to identify clusters but also to convey their inherent relationships and structures graphically. This task finds applications across various domains, from exploratory data analysis to pattern recognition, where understanding the inherent organization of data fosters insights and informed decision-making \citep{luxburg2012clustering}. Effective visualization-driven clustering strategies contribute to the development of interpretable and actionable representations \citep{assent2012clustering}, facilitating a deeper understanding of intricate dataset structures, even in high-dimensional data \citep{strehl2003relationship}. 

\subsection{Problem Space Design}
We have synthesized 6.130 datasets to represent a vast range of clustering problems to compose the $\mathcal{P}(X)$. These datasets were created by combining  different ranges of important clustering characteristics \citep{zellinger2023repliclust}, namely: number of dimensions, number of clusters, quantity of data points, imbalance ratio as well as the shapes of the clusters, described in \autoref{tab:problem_space}. 

\begin{table}[!htp]
    \centering
    \caption{Dataset synthesized for the PoAC's \textit{problem space} described in \autoref{sec:setup}.}
    \begin{tabular}{p{4cm}c}
    \toprule
        \textbf{Feature} & \textbf{Range} \\
        \midrule
         Dimensions & 2 - 100 \\
         Clusters &  2 - 35 \\
         Samples & 150 - 5000 \\
         Overlap & 1e-6 - 1e-5 \\
         Aspect Ref & 1.5 - 5 \\
         Aspect Max Min & 1 - 5 \\
         Radius Max Min & 1 - 5 \\
         Distributions & normal - exponential - gumbel \\
         Imbalance Ratio & 1 - 3 \\
    \bottomrule
    \end{tabular}
    \label{tab:problem_space}
\end{table}

To compose the observations $X$, we conducted a data augmentation process to generate a range of different partitionings of each dataset in $\mathcal{P}(X)$. This augmentation process consisted of inserting Gaussian noise \citep{lopes2019improving} one hundred times into the cluster labels of the $\mathcal{P}(X)$ datasets. The observations $X$ are mapped in terms of $\mu$, which consists of (i) the unsupervised meta-features extracted from the original datasets and CVIs calculated on the augmented partitioned data of $\mathcal{P}(X)$, and (ii) the external CVI ($m$), used as the target for the surrogate model.

\subsection{Feature Space Mapping}\label{app:metrics}

\subsubsection{Dataset Meta-features}
The meta-features selected for this work were chosen based on \citep{lorena2019how} given the comprehensibility of the proposed set of descriptors. In summary the relevant ones in the context of this work amount to 37, and are described in \autoref{tab:app-meta-features}.

\begin{table}[!htp]
    \centering
    \caption{List of meta-features extracted from datasets to compose the surrogate model.}
    \begin{tabular}{lc>{\raggedright\arraybackslash}m{7cm}}  
        \toprule
        \textbf{Meta-Features} & \textbf{Group} & \textbf{Description} \\
        \midrule
        attr\_conc & info-theory & Concentration coefficient of each pair of distinct attributes. \\
        attr\_ent & info-theory & Shannon’s entropy for each predictive attribute. \\ 
        attr\_to\_inst & general & Ratio between the number of attributes. \\
        cohesiveness & concept & Improved version of the weighted distance, that captures how dense or sparse is the example distribution. \\
        cov & statistical & Absolute value of the covariance of distinct dataset attribute pairs. \\
        eigenvalues & statistical & Eigenvalues of covariance matrix from dataset. \\
        inst\_to\_attr & general & Ratio between the number of instances and attributes. \\
        iq\_range & statistical & Interquartile range (IQR) of each attribute. \\
        mad & statistical & Median Absolute Deviation (MAD) adjusted by a factor. \\
        median & statistical & Median value from each attribute. \\
        nr\_attr & general & Total number of attributes. \\
        nr\_cor\_attr & statistical & Number of distinct highly correlated pair of attributes. \\
        nr\_inst & general & Number of instances (rows) in the dataset. \\
        one\_itemset & itemset & One itemset meta-feature. \\
        sd & statistical & standard deviation of each attribute. \\
        sparsity & statistical & Calculates the (possibly normalized) sparsity metric for each attribute. \\
        t2 & complexity & Average number of features per dimension. \\
        t3 & complexity & Average number of PCA dimensions per points. \\
        t4 & complexity & Ratio of the PCA dimension to the original dimension. \\
        t\_mean & statistical & Trimmed mean of each attribute. \\
        two\_itemset & itemset & Two itemset meta-feature. \\
        var & statistical & Variance of each attribute. \\
        wg\_dist & concept & Weighted distance, that captures how dense or sparse is the example distribution. \\
        \bottomrule
    \end{tabular}
    
    \label{tab:app-meta-features}
\end{table}

\subsubsection{Internal CVI}
We opted to use Silhouette Index (SIL) \citep{rousseeuw1987silhouettes} and Davies-Bouldin Score (DBS) \citep{davies1979cluster} as CVIs. Their selection as visualization descriptors for clustering problems is grounded in their ability to provide comprehensive insights into the quality and distinctiveness of clustering solutions. A high SIL suggests visually distinct and well-separated clusters, enhancing the interpretability of visual representations \citep{Shahapure2020ClusterQA, BAGIROV2023109144}. Complementing this, a low DBS reinforces the notion of visually cohesive and distinguishable clusters, contributing to an enhanced visual understanding \citep{Maulik2002PerformanceEO,Thomas2013NewVO}. The two CVIs assess complementary aspects in regards to visualization, their combination allows for a more reliable optimization function, i.e., providing more sound clustering solutions. We chose not to include other internal CVIs due to (i) not being complementary (i.e., measuring the same facets of the problem) and (ii) including many dimensions to optimize makes the optimization problem more challenging and resource costly.

SIL ranges from -1 to 1: a value close to 1 for a data point indicates that it is well-clustered, meaning it is further away from neighboring clusters than its own; a value around 0 suggests that the point lies on the boundary between two clusters; and a negative value indicates that the point may be assigned to the wrong cluster, as it is closer to a different cluster than its own. The average SIL across all points in a dataset provides an overall measure of clustering quality \citep{rousseeuw1987silhouettes}. Let the \(i, j\) be n-dimensional feature vectors (data points) and \(i, j\) $\in C_{I}$; $C$ as a given cluster of the dataset $D$; \(d(i, j)\) as the average Euclidean distance between data points $i$ and $j$; \(\lvert C_I \rvert\) as the number of data points in cluster \(C_{I}\) and \( |N| \) is the total number of data points present in the dataset $D$. The Silhouete Score for \( i \) can be calculated using Equation \eqref{eq:sil-single} if and only if \(|C_{I}| > 1\).

\begin{equation}\label{eq:sil-single}
    \mathcal{SIL}(i) = \frac{b(i) - a(i)}{\max\{a(i), b(i)\}}
\end{equation}

Where, \( a(i) \) is the average distance from the \( i \)-th data point to the other data points in the same cluster, given by Equation \eqref{eq:a(i)}.
    
    \begin{equation}\label{eq:a(i)}
        a(i) = \frac{1}{\lvert C_I \rvert - 1} \sum_{j \in C_I, j \neq i} d(i, j)
    \end{equation}

And, $b(i)$ is the smallest average distance between the $i$-th data point of $C_{I}$ and the $k$-th data point of $C_{K}$, where $C_{I}$ $\neq$ $C_{K}$, given by Equation \eqref{eq:b(i)}.

    \begin{equation}\label{eq:b(i)}
        b(i) = \min_{K \neq I} \frac{1}{\lvert C_{K} \rvert} \sum_{k \in C_{K}} d(i, k) 
    \end{equation}

Finally, the overall SIL for the entire dataset $D$ is the average of the SIL for all data points, as given by Equation \eqref{eq:sil-overall}:

    \begin{equation}\label{eq:sil-overall}
        \mathcal{SIL}_{D} = \frac{1}{|N|} \sum_{i=1}^{N} \mathcal{SIL}(i)
    \end{equation}
    
The DBS, as introduced by \citep{davies1979cluster}, is defined as the ratio of compactness within clusters and the separation between clusters. It ranges from 0 to infinite, where a DBS closer to 0 indicates better (more compact) clustering, with well-defined and separated clusters. Let \(C_I\) be a cluster of a dataset $D$; $\sigma$ the average Euclidean distance from each data point of $C_I$ to its centroid $A_I$; \( d(C_I, C_K) \) be the Euclidean distance between the $A_I$ and $A_K$; and the data-point $i \in C_I$; $|C|$ be the total number of clusters in $D$; and $|C_I|$ be the total number of data points in $C_I$. 

Where, the compactness of $C_I$ is given by Equation \eqref{eq:sigma}:
\begin{equation}\label{eq:sigma}
    \sigma_I = \frac{1}{\lvert C_I \rvert} \sum_{i = 1}^{|C_I|} || i - A_I||
\end{equation}

The distance between the centroids is given by Equation \eqref{eq:distance}:
\begin{equation}\label{eq:distance}
    d(C_I, C_K) = ||A_I - A_K||
\end{equation}

And, the DBS for the entire dataset $D$ is calculated using the Equation \eqref{eq:dbi}:
\begin{equation}\label{eq:dbi}
    \mathcal{DBS}_D = \frac{1}{|C|} \sum_{I=1}^{|C|} \max_{K\neq I} \left( \frac{\sigma_I + \sigma_K}{d(C_I, C_K)} \right)
\end{equation}

\subsection{Surrogate Modeling}
\subsubsection{External CVI}
For $m$, we chose the Adjusted Random Index, as defined in \autoref{eq:a(i)}, to quantify the difference between the augmented partitions and the ground truth according to the original dataset. The intent for the composition of $\mu$ and $m$ in this way is to create a range of possible partitionings that represents how visually different they are to the original clustering. By jointly employing these metrics with the meta-features, we aim to capture a nuanced understanding of the visual separability and coherence of clusters in our evaluation, ensuring a robust assessment of the visibility of clustering patterns in diverse datasets.

The Adjusted Random Index (ARI) is a essentially an expansion of the Rand Index (RI), which is a measure of similarity between two partitionings \citep{hubert1985comparing}. It is an external CVI, meaning it is useful for evaluating the performance of clustering algorithms when the ground truth is known. The RI considers pairs of samples and classifies them as either concordant or discordant based on whether they are placed in the same or different clusters in both partitionings. It ranges from 0 to 1, where 1 indicates perfect agreement between the two partitionings, and 0 indicates no agreement beyond that expected by chance.

Let \( n \) be the number of elements in the set \( D = \{o_1, \dots, o_n\} \), and let \( U = \{U_1, \dots, U_I\} \) and \( V = \{V_1, \dots, V_J\} \) represent two different partitionings of \( D \) into \( I \) and \( J \) clusters, respectively.

The RI is calculated using the formula \eqref{eq:rand-ind}:

\begin{equation}\label{eq:rand-ind}
RI = \frac{\alpha + \beta}{\binom{n}{2}}
\end{equation}

Where:
\begin{itemize}
    \item \( \alpha \) is the number of pairs of elements \( (o_i, o_j) \) that are placed in the \textbf{same cluster} in both \( U \) and \( V \).
    \item \( \beta \) is the number of pairs of elements \( (o_i, o_j) \) that are placed in \textbf{different clusters} in both \( U \) and \( V \).
\end{itemize}

For all \( 1 \leq i, j \leq n, i \neq j \):

\begin{equation}\label{eq:alfa}
    \alpha = \left| \{(o_i, o_j) \mid o_i, o_j \in U_k \text{ and } o_i, o_j \in V_l \text{ for some } k, l\} \right|
\end{equation}

\begin{equation}\label{eq:beta}
\begin{split}
    \beta = \left| \{(o_i, o_j) \mid o_i \in U_{k_1}, o_j \in U_{k_2} \text{ and } o_i \in V_{l_1}, o_j \in V_{l_2}, \right. & \\
    \left. \text{ for } k_1 \neq k_2 \text{ and } l_1 \neq l_2 \} \right|
\end{split}
\end{equation}

In this sense, \( \alpha \) counts the number of pairs in the same cluster in both partitionings, while \( \beta \) counts pairs in different clusters in both partitionings. The denominator \( \binom{n}{2} \) represents the total number of possible pairs of elements in \( D \).

The ARI accounts for chance agreement, providing a normalized measure that ranges from -1 to 1. A score of 1 indicates perfect agreement, 0 suggests agreement expected by chance, and negative values imply worse-than-chance agreement. The formula for the ARI is given by:

\begin{equation}
    \mathcal{ARI} = \frac{ \sum_{ij} \binom{n_{ij}}{2} - [\sum_i \binom{\alpha_i}{2} \sum_j \binom{\beta_j}{2}] / \binom{n}{2} }{ \frac{1}{2} [\sum_i \binom{\alpha_i}{2} + \sum_j \binom{\beta_j}{2}] - [\sum_i \binom{\alpha_i}{2} \sum_j \binom{\beta_j}{2}] / \binom{n}{2} }
\end{equation}

This formula normalizes the RI by accounting for random chance, making the ARI a valuable tool for assessing clustering accuracy in various fields, including biology, image analysis, and social sciences.

\subsubsection{Surrogate Model}
As the surrogate model $f_m$, we trained a Random Forest regressor \citep{breiman2001random} considering the observations $X$ and the independent variables $\mu$, while also considering $m$ as the dependent variable for the regression. The adoption of Random Forest in this study stems from its versatility and robustness in capturing complex relationships within data \citep{hastie2009random}, making it well-suited for modeling the intricate nuances of clustering solutions. Additionally, the feature importance derived from the Random Forest model is relevant for its ability to provide insights into which variables significantly influence the clustering performance. By doing so, the surrogate model offers a reliable fitness function for the PS, effectively guiding the optimization process.

\begin{figure}[!htp]
    \centering
    \includegraphics[width=0.7\textwidth]{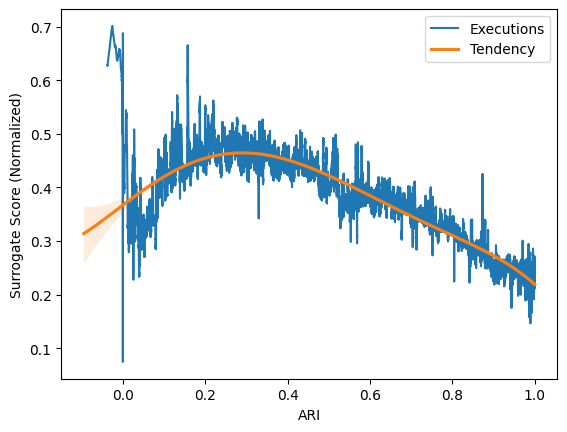}
    \caption{Correlation between ARI and surrogate model.}
    \centering
    \label{fig:correlation}
\end{figure}

During the training, the Random Forest model was tuned to handle the diverse range of input features, capturing the complex relationships between meta-features, CVIs, and the clustering outcomes. To ensure robust model performance, we employed 10-fold cross-validation. To assess the trained model's performance in predicting the ARI based on the given meta-features and CVIs, we used validation metrics, such as the mean squared error (MSE) and R-squared (R²) score. The model obtained a R² of 0.7 and a MSE of 0.2, indicating a good fit, ensuring that the Random Forest model was neither overfitting nor underfitting the data, as shown in \autoref{fig:correlation}. 

\begin{figure}[!htp]
    \centering
    \includegraphics[width=1\textwidth]{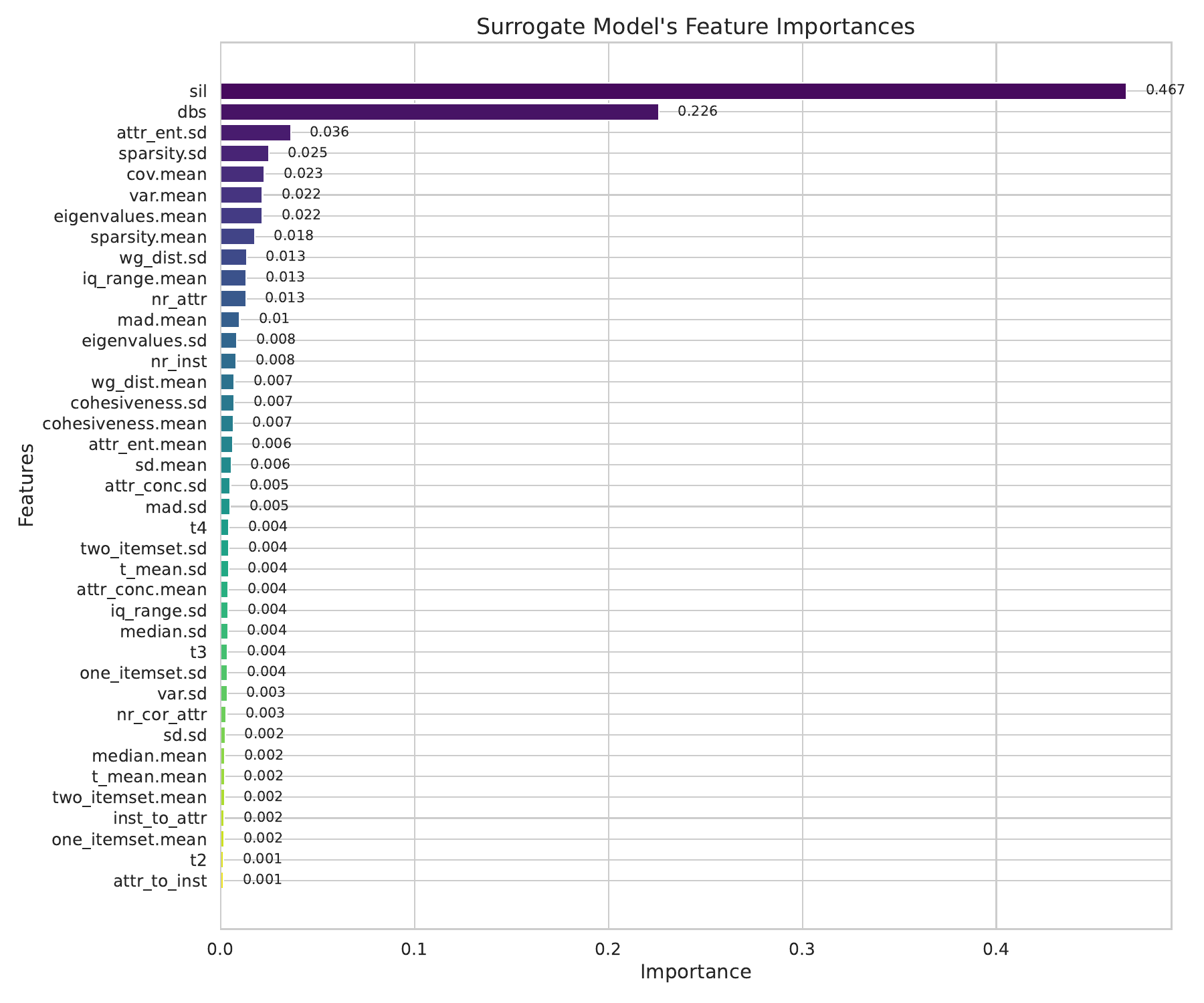}
    \centering
    \caption{Surrogate model's Feature Importance.}
    \label{fig:feature_importance}
\end{figure}

According to the feature importance plot of the surrogate model, as displayed in \autoref{fig:feature_importance}, the most critical feature of the trained Random Forest is the SIL score, with a substantial importance score of 0.467, indicating that it plays a dominant role in the model's decisions. Following this, the DBS score also holds considerable importance with a score of 0.22. These two CVIs suggest that the model heavily relies on cluster compactness and separation measures when predicting the performance of clustering algorithms, which of course is useful in the context of clustering for visualization. This can be a consequence of SIL and DBS having a much higher variation in the meta-database since the augmented datasets will always have the same meta-features.

On the other hand, the statistical and information-theoretic features such as `attr\_ent.sd` (Shannon’s entropy standard deviation), `sparsity.sd` (sparsity standard deviation), `cov.mean` (mean covariance), and `var.mean` (mean variance) have lower importance scores, ranging from 0.03 to 0.02. These meta-features, while contributing to the model's decisions, do so to a much lesser extent than the leading CVIs. This indicates that while the randomness or variability in entropy, sparsity, and covariance definitely plays a role, it is not as pivotal as the CVIs.

Furthermore, many other statistical measures, including median.mean (average median), `t\_mean.sd` (standard deviation of trimmed mean), and `attr\_to\_inst` (the ratio of attributes to instances), show very low importance scores, generally below 0.01. This implies that the model does not heavily rely on these specific statistical measures or general dataset characteristics when predicting clustering performance.

In summary, the surrogate model's reliance on CVIs like the SIL and DBS score highlights the critical role of these indices in evaluating clustering quality, while the statistical and information-theoretic features contribute less significantly to the model's predictions. This suggests that the model finds more value in the direct assessment of clustering outcomes than in the underlying statistical properties of the datasets. It is important to note though, that these conclusions are relevant only to the trained Random Forest for the visualization problem, and that other clustering problems with different CVIs might present different importances.

\subsection{Function Optimization}
\begin{figure}[!htp]
    \centering
    \includegraphics[width=1\textwidth]{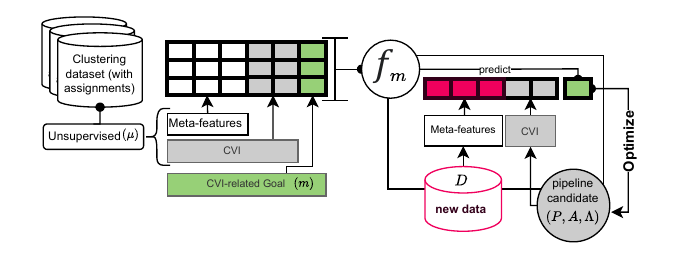}
    \caption{Pipeline optimization process for clustering problems within the PoAC framework. The process begins by extracting meta-features (\(\mu\)) and CVI from clustering datasets. This CVI-related goal (\(m\)) serves as the target for the surrogate model \(f_m\), which is used to predict the quality for pipeline candidates (\(P, A, \Lambda\)) on new, unseen data (\(D\)). The surrogate model then optimizes the pipeline, evaluating potential solutions based on their predicted (\(m\)), using extracted meta-features from the new data and CVI from pipelines to inform the optimization process.}
    \label{fig:opt}
\end{figure}

Lastly, we extended the capabilities of the AutoML framework Tree-based Pipeline Optimization Tool (TPOT), presented by \citep{olson2016tpot}, to allow it to synthesize pipelines for clustering problems. This extension consists of a set of important modifications. Mainly, the incorporation of the surrogate model to serve as a fitness function for the evolutionary optimization process. This entails the extraction of meta-features on new unseen clustering datasets, and the measuring of both SIL and DBS for each synthesized clustering pipeline solution, as displayed in \autoref{fig:opt}. TPOT was chosen due to its robustness, high level of maintenance, and widespread use within the machine learning community, making it a reliable and well-supported tool for extending into new domains. These qualities make TPOT particularly suitable for our purposes, as it provides a stable and flexible framework for automating complex machine learning tasks.

Furthermore, we have replaced the conventional TPOT's classification operators with clustering counterparts, as displayed in \autoref{tab:app-opts}. Through these modifications, we effectively extended TPOT's utility to the realm of clustering problems, providing a versatile and adaptive AutoML tool.

\begin{table}[!htp]
\centering
\caption{List of Operators we added to the configuration of TPOT.}
\begin{tabular}{m{0.20\linewidth} m{0.15\linewidth} m{0.55\linewidth}}
\toprule
\textbf{Operator} & \textbf{Type} & \textbf{Hyperparameters} \\
\midrule
Agglomerative Clustering & cluster &                
\begin{itemize}
    \item \textbf{n\_clusters}: range(2, 23)
    \item \textbf{metric}: euclidean
    \item \textbf{linkage}: ward
\end{itemize}        
\\ 
DBSCAN & cluster &                                  
\begin{itemize}
    \item \textbf{eps}: range(2, 23)
    \item \textbf{min\_samples}: [1e-3, 1e-2, 1e-1, 1., 10., 100.]
    \item \textbf{metric}: [10, 25, 50]
    \item \textbf{leaf\_size}: [3, 5, 10, 15, 20, 25, 30, 35, 40, 45, 50]
\end{itemize}        
\\ 
KMeans & cluster &                                  
\begin{itemize}
    \item \textbf{n\_clusters}: range(2, 23)
    \item \textbf{init}: [k-means++, random]
\end{itemize}        
\\ 
Mini Batch KMeans & cluster &                 
\begin{itemize}
    \item \textbf{n\_clusters}: range(2, 23)
    \item \textbf{eps}: range(2, 23)
\end{itemize}        
\\ 
Spectral Clustering & cluster &                 
\begin{itemize}
    \item \textbf{n\_clusters}: range(2, 23)
    \item \textbf{eigen\_solver}: [arpack, lobpcg, amg]
    \item \textbf{affinity}: [nearest\_neighbors, rbf, precomputed, precomputed\_nearest\_neighbors]
\end{itemize}        
\\ 
MinMax Scaler & preprocessing & -       
\\ 
Normalizer & preprocessing &                 
\begin{itemize}
    \item \textbf{norm}: [l1, l2]
\end{itemize}        
\\ 
Standard Scaler & preprocessing & -     
\\ 
Variance Threshold & feature selection &                 
\begin{itemize}
    \item \textbf{threshold}: [0.1, 0.25]
\end{itemize}        
\\ 
PCA & decomposition &              
\begin{itemize}
    \item \textbf{n\_components}: [2, 3, 5, 10]                  
\end{itemize}        
\\ 
Fast ICA & decomposition &                 
\begin{itemize}
    \item \textbf{n\_components}: [2, 3, 5, 10]                  
\end{itemize}        
\\
\bottomrule
\end{tabular}
\label{tab:app-opts}
\end{table}

\clearpage
\subsection{Baselines}
\label{sec:experiments}

We designed and performed experiments to evaluate the quality of PoAC when compared to the reproducible state-of-the-art frameworks that focus on CASH, namely \textit{ML2DAC} from \citep{treder2023ml2dac}, \textit{Autocluster} proposed by \citep{liu2021autocluster}, \textit{cSmartML} as presented in \citep{elshawi2021csmartml} and, finally, \textit{AutoML4Clust} presented in \citep{tschechlov2021automl4clust}. As for PoAC, we specifically created an instance of the framework for visualization as a clustering problem \citep{ezugwu2022comprehensive} and assessed the performance of pipeline optimizations. In the context of the proposed \textit{problem-oriented} methodology, prioritizing visualization as a clustering problem underscores its significance as a user-defined goal, catering to scenarios where the interpretability and visual clarity of clustering solutions are paramount.

We considered two distinct groups of datasets to comprehensively evaluate the frameworks in the context of clustering. The first group is composed of datasets used in the benchmark work done by \citep{silva2024benchmarking}, it consists of one hundred synthetic clustering datasets with ranging degrees of instances distributions, clusters separations and densities, among others, as described in \autoref{tab:validation-synth}. 

\begin{table}[!htp]
    \centering
    \caption{Dataset validation group for experiments in \autoref{sec:experiments}}
    \begin{tabular}{p{4cm}l} 
        \toprule
        \textbf{Feature} & \textbf{Range} \\
        \midrule
         Dimensions & 2 - 100 \\
         Clusters &  2 - 35 \\
         Samples & 150 - 5000 \\
         Overlap & 1e-6 - 1e-5 \\
         Aspect Ref & 1 - 10 \\
         Aspect Max Min & 1 - 10 \\
         Radius Max Min & 1 - 10 \\
         Imbalance Ratio & 1 - 3 \\
    \bottomrule
    \end{tabular}
    \label{tab:validation-synth}
\end{table}

The second group (\autoref{table:app-real-world}), encompassed twenty two datasets sourced from the UCI repository \citep{uci@2017}, providing a real-world dimension to our analysis, they were chosen for their usage in the validation of related works, namely: Ml2dac and AutoCluster.
\clearpage

\begin{table}[!htp]
    \centering
\caption{Real-world datasets validation group.}
\begin{tabular}{lrrr}
    \toprule
    \textbf{Dataset} &  \textbf{Instances} &  \textbf{Num Features} &  \textbf{Number of Clusters} \\
    \midrule
   arrhythmia &        452 &           262 &             13  \\ 
balance-scale &        625 &             4 &              3  \\ 
  dermatology &        366 &            34 &              6  \\ 
        ecoli &        336 &             7 &              8  \\ 
       german &       1000 &             7 &              2  \\ 
        glass &        214 &             9 &              6  \\ 
     haberman &        306 &             3 &              2  \\ 
heart-statlog &        270 &            13 &              2  \\ 
         iono &        351 &            34 &              2  \\ 
         iris &        150 &             4 &              3  \\ 
       letter &      20000 &            16 &             26  \\ 
      segment &       2310 &            19 &              7  \\ 
        sonar &        208 &            60 &              2  \\ 
          tae &        151 &             5 &              3  \\ 
          thy &        215 &             5 &              3  \\ 
      vehicle &        846 &            18 &              4  \\ 
        vowel &        990 &            10 &             11  \\ 
         wdbc &        569 &            30 &              2  \\ 
         wine &        178 &            13 &              3  \\ 
         wisc &        699 &             9 &              2  \\ 
        yeast &       1484 &             8 &             10  \\ 
          zoo &        101 &            16 &              7  \\ 
\bottomrule
          
\end{tabular}
\label{table:app-real-world}
\end{table}

By incorporating both synthetic and real-world datasets, our experimental design aimed to provide a holistic assessment of the PoAC's performance across a spectrum of clustering challenges, thereby enhancing the generalizability and applicability of our findings.
\section{Results and Discussion}

We applied each one of the frameworks, as well as the proposed PoAC, on the group of one hundred validation datasets. The results are presented in \autoref{table:synth_mean_ari_sil_dbs}. The obtained results pertaining ARI will be discussed in \autoref{sec72}.

\begin{table}[!htb]
\centering
\caption{Frameworks performance regarding ARI, SIL and DBS for the validation datasets.}
\begin{tabular}{lrrr}
\toprule
                 \textbf{Framework} &  \textbf{Mean ARI} &  \textbf{Mean SIL} &  \textbf{Mean DBS} \\
\midrule
        \textbf{PoAC} &   \textbf{ 0.70} &    \textbf{0.54} &    \textbf{0.76} \\
              ML2DAC &    0.67 &    0.39 &    1.67 \\
         Autocluster &    0.59 &    0.23 &    2.42 \\
        AutoML4Clust &    0.58 &    0.36 &    1.39 \\
            cSmartML &    0.38 &    0.24 &    1.37 \\
\bottomrule
\end{tabular}
\label{table:synth_mean_ari_sil_dbs}
\end{table}

PoAC exhibited outstanding results, obtaining an average SIL of 0.54, suggesting well-defined and separated clusters, surpassing other frameworks in this regard. ML2DAC, AutoML4Clust, followed with a SIL of 0.39 and 0.36, respectively, while cSmartML and Autocluster scored lower at 0.24 and 0.23, respectively. Moreover, in terms of DBS, PoAC again outperformed other frameworks, achieving a favorable score of 0.76, indicative of compact and well-separated clusters. Following PoAC, cSmartML obtained a score of 1.37, AutoML4Clust recorded 1.39, ML2DAC reached 1.67, and Autocluster received a score of 2.42. These results show that PoAC is able to maintain output quality by balancing the analysis dimensions. The other frameworks do not maintain their position in the ranking, meaning that they attempt to maximize only one aspect of the visualization problem, leading to poor clustering solutions. Moreover, PoAC's prowess in synthesizing pipelines that yield visually cohesive and well-defined clusters is demonstrated, showcasing its effectiveness in the context of clustering for visualization tasks. We interpret PoAC's excellence in this task due to its correlation between clustering goal and optimization dimensions, a unique feature of the framework when compared to competitors.

\begin{figure}[!htp]
    \centering
    \includegraphics[width=1\textwidth]{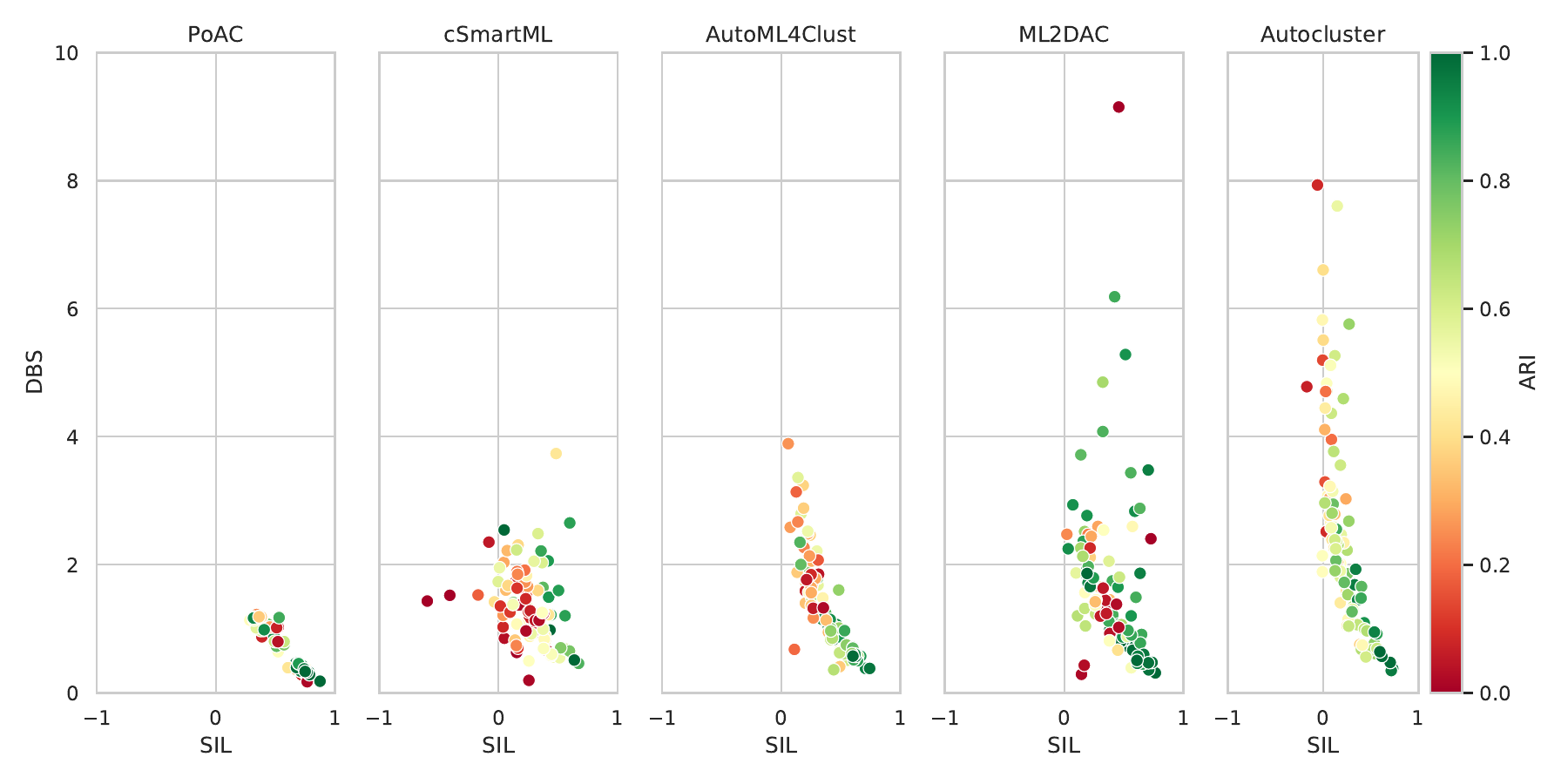}
    \centering
    \caption{Frameworks sorted ascendingly by DBS' variance.}
    \label{fig:sil_dbs_comparison}
\end{figure}

The impact of the difference in the clusterings solutions generated is demonstrated in \autoref{fig:sil_dbs_comparison}, which assess the SIL and DBS values for the solutions of each AutoML framework for every dataset in the first validation group. It also displays the level of ARI measured per clustering. According to this analysis, the PoAC's solutions are concentrated, with the lowest variance, inside the intersection of optimal values for SIL and DBS, while also presenting the highest ARI in the group of frameworks.

\subsection{Clustering Performance Analysis}
\label{sec72}
\begin{figure}[!htp]
    \centering
    \includegraphics[width=1\textwidth]{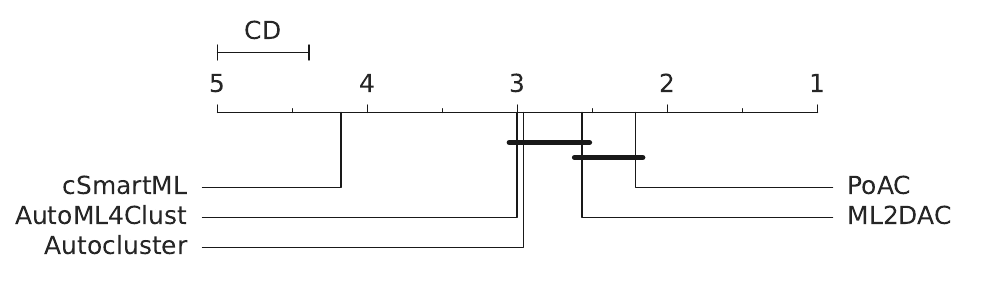}
    \caption{Populations ranked by Mean Rank according to the Nemenyi's test on ARI, where the smallest value indicates the leading population. From last, to first: cSmartML (MR=4.18), AutoML4Clust (MR=3.01), Autocluster (MR=2.96), ML2DAC (MR=2.57), and PoAC (MR=2.22).}
    \label{fig:nemenyi-ari}
\end{figure}

When comparing the similarities between the solutions from the evaluated frameworks and the original clusters for the dataset validation group, the results unveiled a noteworthy performance. Our proposed PoAC method achieved a mean ARI of 0.70. ML2DAC secured the second position with a 0.67 mean ARI, followed by AutoML4Clust at 0.59, Autocluster at 0.58, and cSmartML at 0.38. We conducted a statistical analysis for 5 populations with 100 paired samples of ARI values. The first analysis showed that some of the populations did not present normal distributions, in particular the populations of AutoML4Clust, Autocluster, and PoAC. Since we have a set of more than two populations and some of them are not normal, we applied the non-parametric Friedman test with a confidence level of 95\% that determined that there is a significant difference between the median values of the populations. Based on the post-hoc Nemenyi test \citep{nemenyi1963distribution}, with a critical distance (CD) of 0.61, as shown in \autoref{fig:nemenyi-ari}, we can affirm that there are no significant differences within the following groups: AutoML4Clust, Autocluster, and ML2DAC; ML2DAC and PoAC. All other differences are significant, namely cSmartML when compared to any other population and PoAC when compared to AutoML4Clust and Autocluster.

To better assess the differences in capabilities between ML2DAC and PoAC in generating clustering solutions, we replicated the experiments conducted in the ML2DAC paper \citep{treder2023ml2dac} by evaluating the two frameworks using the datasets from the second validation group, thereby enabling a direct comparison with our proposed approach using real-world scenarios. The configuration identified by ML2DAC achieved a maximum ARI of 0.27, with their simplest configuration attaining 0.19. In comparison, our proposed method exhibited a competitive performance, achieving an ARI of 0.22.

The comparison of our proposed method with other existing frameworks reveals distinctive advantages. In the first experiment, PoAC exhibited the best performance showcasing the efficacy of using a surrogate model as an fitness function for PS in clustering. Furthermore, an important aspect is the adaptability of the surrogate model trained by PoAC on clustering solutions with a specific emphasis on the visualization goal. Despite its specialization, the surrogate model demonstrated a comparable performance to ML2DAC when applied to datasets not specifically clustered for visualization, as evident in the UCI group. Remarkably, PoAC achieved this level of performance without requiring any additional fine-tuning or configuration adjustments. This versatility showcases the robustness of our methodology, emphasizing its potential across diverse clustering problems without the need for intricate parameter adjustments. 

\subsection{Generated Pipeline Complexity}
\label{sec73}
\begin{figure}[!htp]
    \centering
    \includegraphics[width=1\textwidth]{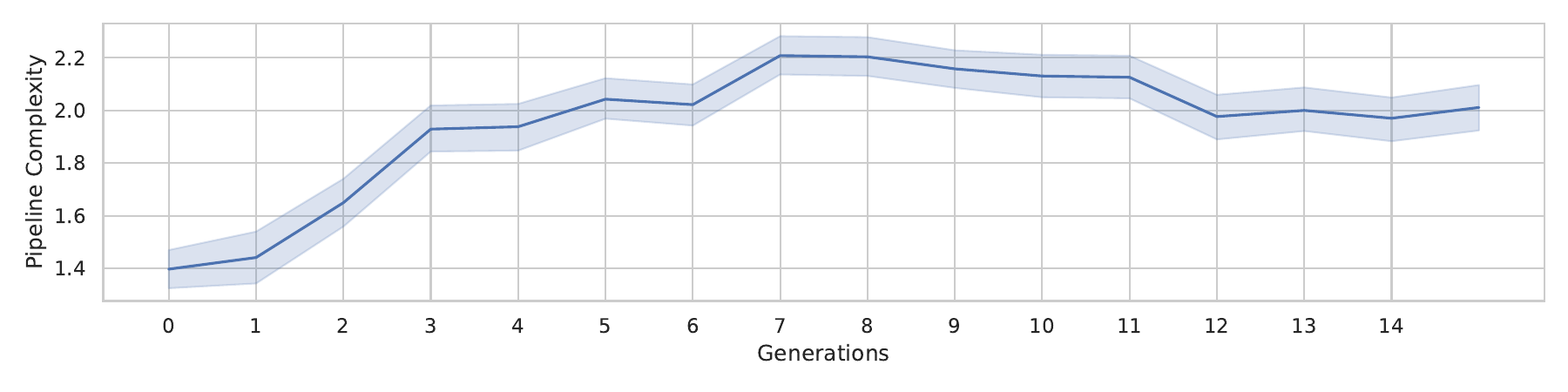}
    \caption{Recommended pipelines' complexity throughout the evolutionary process.}
    \label{fig:pipeline_complexity_evolution}
\end{figure}

Through an analysis of the mean pipeline complexity (i.e., number of pipeline steps) across generations, our findings reveal a notable trend within the PoAC framework. Over successive generations, PoAC exhibits a consistent evolution towards pipelines with a mean complexity of at least two steps, as shown in \autoref{fig:pipeline_complexity_evolution}. The stabilization around a complexity of two steps signifies an optimal balance between the incorporation of essential preprocessing actions and a clustering algorithm. This trend not only underscores the adaptability of PoAC but also highlights its efficiency in autonomously converging towards suitable.

\begin{figure}[!htp]
    \centering
    \includegraphics[width=1\textwidth]{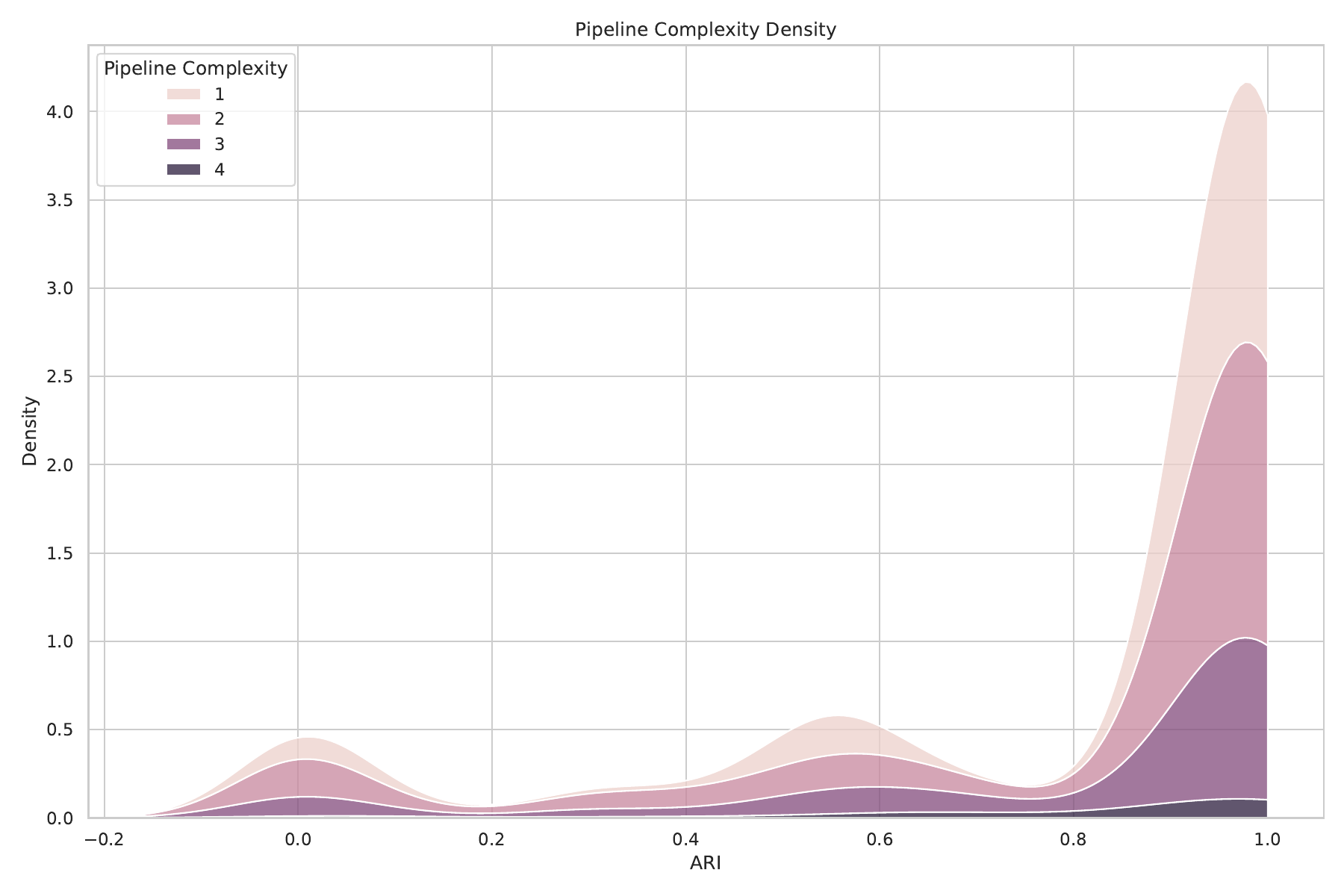}
    \caption{Recommended pipelines' performance and complexity}
    \label{fig:pipeline_complexity}
\end{figure}

As depicted in \autoref{fig:pipeline_complexity}, the density of pipeline complexity values shows a clear differentiation based on ARI values. Pipelines with lower ARI scores are associated with a somewhat even distribution of complexities, reflecting a more exploratory nature in the search for optimal solutions. Conversely, pipelines achieving higher ARI values tend to cluster around a complexity of one and two steps, aligning with the observed results in \autoref{fig:pipeline_complexity_evolution}. However, it is important to note that there is a small, but not insignificant volume of pipelines with three and four steps. This reinforces the notion that within the PoAC framework, the evolution towards simpler pipelines is not at the expense of performance, but rather a convergence towards the most efficient configurations for the task at hand.

\begin{figure}[!htp]
    \centering
    \includegraphics[width=1\textwidth]{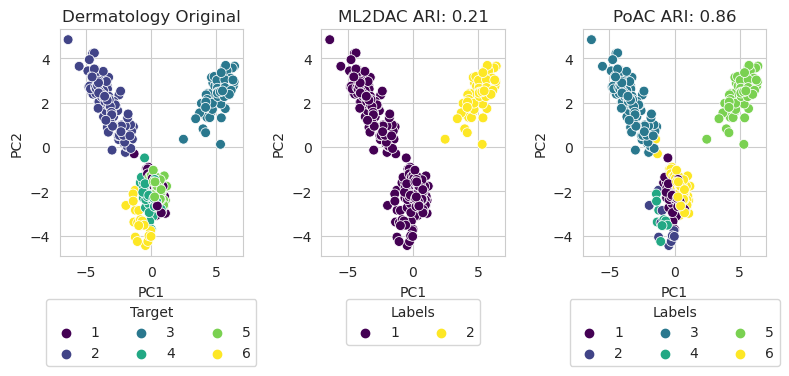}
    \caption{Dermatology showcase example.}
    \label{fig:showcase}
\end{figure}

In the realm of clustering problems, the choice and sequence of ML techniques can significantly impact performance. \autoref{fig:showcase} illustrates this phenomenon in the context of the UCI dataset \textit{dermatology} \citep{misc_dermatology_33}, where two distinct clustering solutions are showcased using a Principal Component Analysis (PCA) plot. PCA is commonly used in clustering to reduce high-dimensional data to two or three dimensions for visualization, helping to reveal patterns or separations between clusters. By projecting data onto a lower-dimensional space, PCA aids in assessing how well clusters are formed, though it may miss non-linear structures \citep{jolliffe2016principal}. The first solution given by ML2DAC solves the CASH task by recommending the \textit{Kmeans} algorithm using 2 as the number of clusters hyperparameter.
PoAC is capable of generating a pipeline for the dermatology dataset with multiple steps: (i) \textit{MinMaxScaler} as a preprocessing step; (ii) \textit{MiniBatchKMeans} as the clustering algorithm; (iii) HPO by setting the batch size to 10 and the number of clusters to 6. Notably, the more complex pipeline presented a more faithful clustering regarding the original one, underscoring the importance of solving the complete pipeline (i.e., PS) instead of only limiting to choose the algorithm and hyperparameters (i.e., CASH).
This empirical demonstration highlights the potential benefits of incorporating automated PS techniques, such as those offered by PoAC, to elevate clustering performance.

%
An inherent strength of the PoAC framework lies in its algorithm-agnostic nature, which sets it apart from other AutoML techniques that rely on predefined clustering algorithms and hyperparameter configurations to build a meta-database. In contrast, PoAC introduces a novel augmentation step during the meta-learning phase, where noise is systematically added directly to the cluster labels of synthetic datasets, rather than depending on a limited set of clustering algorithms and hyperparameters, as is common in related work. This approach allows the surrogate model to learn patterns of partitioning that are less tied to specific algorithms, adding flexibility not only during the training phase but also in the optimization phase. This capability allows PoAC to remain adaptable during the optimization process, as the search space can be dynamically altered without requiring re-training of the surrogate model. As a result, the framework is able to optimize over a broader, more flexible space of potential solutions, further reinforcing its algorithm-agnostic stance.

\subsection{Ablation Study}
It is crucial to ensure that the performance/complexity trade-off is justified. To this end, we conduct an ablation study to evaluate the efficacy of the complete PoAC method against sub-configurations of its components in the task of clustering PS. We named the different optimization strategies considered in this study respectively: 

\begin{enumerate}
    \item \textbf{PoAC SIL} optimizes pipelines based solely on their SIL score, favoring those that demonstrate higher clustering quality as indicated by this metric.
    
    \item \textbf{PoAC DBS} focuses exclusively on the DBS (Dunn's Boundary Separation) score, selecting pipelines that achieve superior performance according to this criterion.
    
    \item \textbf{PoAC CVI} adopts a different approach by creating a surrogate model that combines SIL and DBS scores, specifically examining their non-linear correlation without incorporating the meta-features group. This surrogate model, trained using a Random Forest on the same dataset as the complete PoAC method, provides a more nuanced understanding of the relationship between SIL and DBS.
\end{enumerate}

We ran each of the four optimization strategies ten times for every dataset in the first validation group (\autoref{tab:validation-synth}), and recorded the mean ARI per dataset for each of the strategies.

\begin{figure}[!htp]
    \centering
    \includegraphics[width=0.8\textwidth]{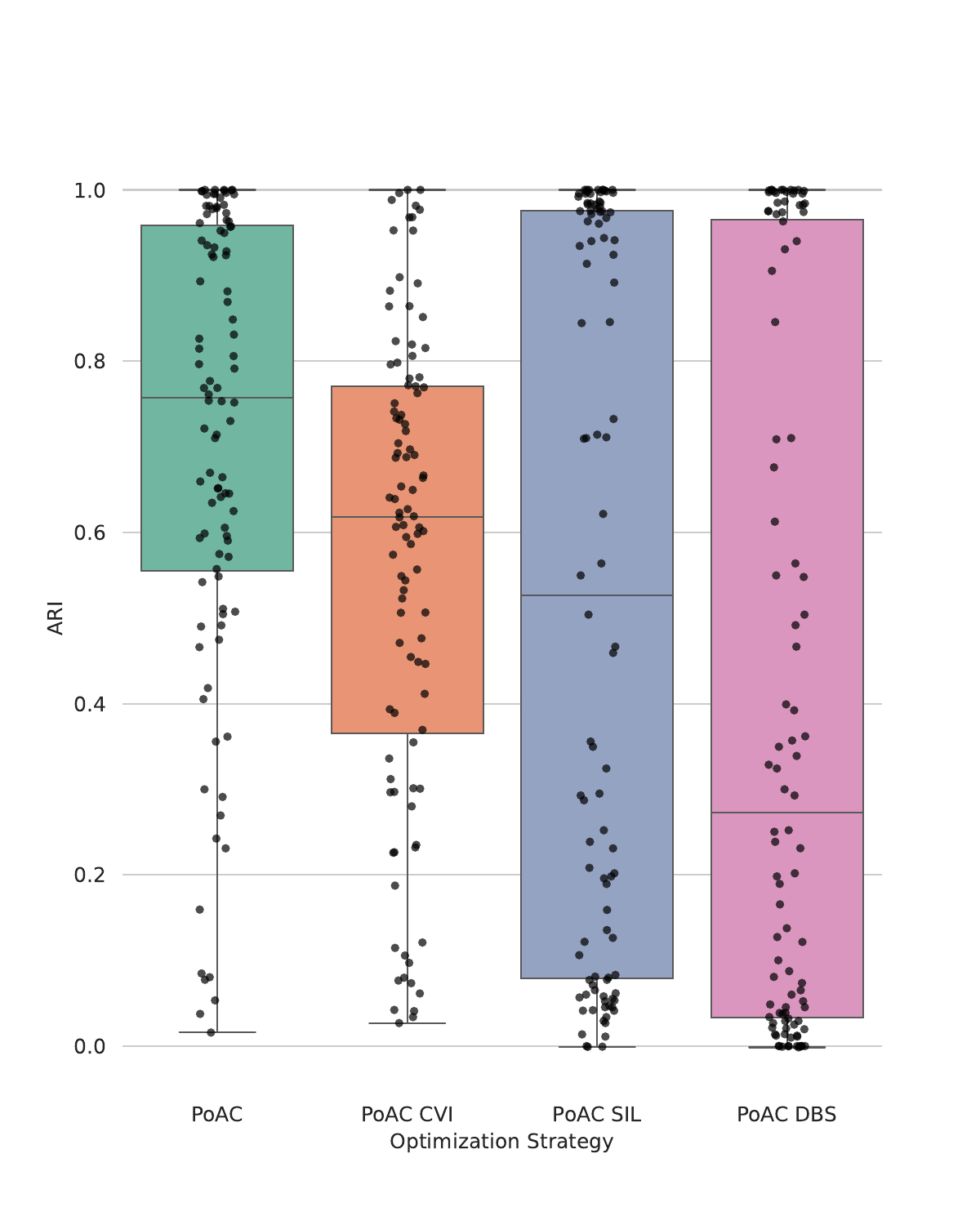}
    \caption{ARI distribution per optimization strategies, namely: Complete \textbf{PoAC} (meta-features+CVI+surrogate model), the \textbf{PoAC CVI} (CVI+surrogate model), \textbf{PoAC SIL} (SIL optimization), \textbf{PoAC DBI} (DBI optimization)}
    \label{fig:ablatiion-boxplot}
\end{figure}

The boxplot analysis in \autoref{fig:ablatiion-boxplot} reveals distinct performance characteristics across the four aforementioned optimization strategies, based on their ARI scores. PoAC presents a relatively narrow interquartile range (IQR), indicating less variability in clustering performance. The upper quartile of the data extends close to the maximum possible ARI score of 1.0, and numerous individual data points are clustered near this upper bound. The median ARI is significantly higher than the other strategies, around 0.8. This suggests that PoAC frequently achieves highly accurate clustering results. While the lower whisker does extend down to approximately 0.1, highlighting some instances of lower performance, PoAC generally offers robust and dependable clustering outcomes.

PoAC CVI, while somewhat similar to PoAC, exhibits a lower median ARI score, around 0.6, and a slightly wider IQR. This wider range might indicate more variability in performance. The lower whisker reaches down to 0.0, meaning that PoAC CVI occasionally produces very poor clustering results. Nevertheless, it also achieves a significant number of high ARI scores, demonstrating that while it may be less consistent than PoAC, it can still be effective in certain cases.

PoAC SIL, shows a wide spread of ARI values. The median ARI is around 0.5, which suggests moderate clustering performance. The IQR is substantial, indicating variability in the results. While it can achieve results around the maximum value, suggesting some instances of excellent performance, the overall inconsistency is notable.

Lastly, the ARI values for PoAC DBS are also widely distributed, covering the full range from 0.0 to 1.0. However, the median ARI is lower than PoAC SIL, and the larger IQR reflects even greater variability in clustering performance, it is in fact the largest in the whole group of strategies. This indicates that while PoAC DBS can occasionally achieve high ARI values, its performance is highly inconsistent, leading to a broad range of outcomes.

\begin{figure}[!htp]
    \centering
    \includegraphics[width=0.8\textwidth]{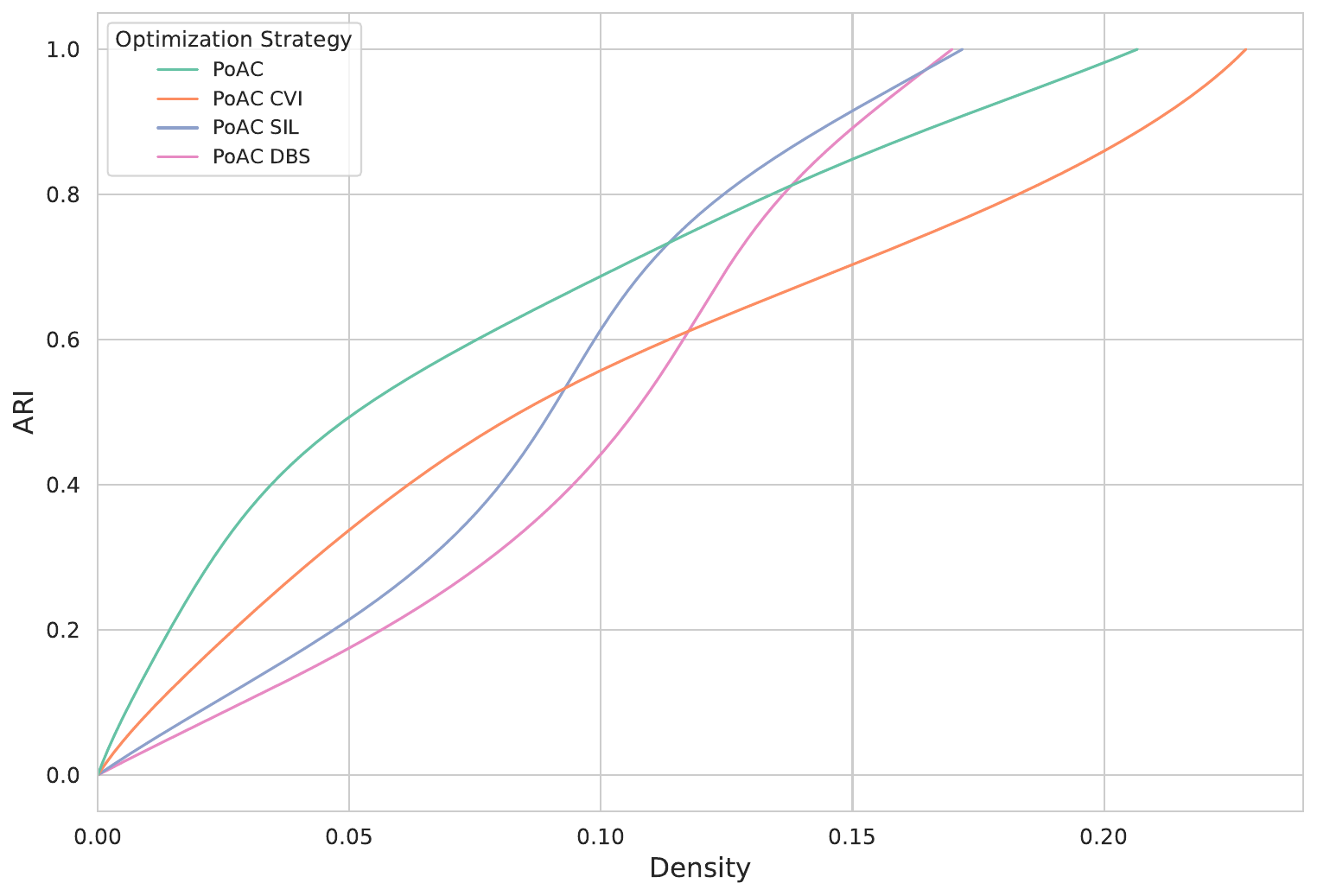}
    \caption{Density Distribution of ARI Scores Across Different Optimization Strategies}
    \label{fig:ablatiion-kde}
\end{figure}

As shown in \autoref{fig:ablatiion-kde}, the optimization strategies that rely on a single CVI tend to have a higher concentration of lower ARI values when compared to the complete and the ablated PoAC. The green line, representing PoAC, has the steepest and most direct rise, indicating that this strategy achieves higher ARI scores more consistently. Most of the density for PoAC is concentrated above the 0.8 ARI mark, suggesting that it tends to produce highly accurate clustering outcomes. The orange line, representing PoAC CVI, also shows a steady rise, though it lags behind PoAC. This indicates that while PoAC CVI is generally effective, it produces slightly lower ARI scores on average compared to PoAC.

The blue (PoAC SIL) and pink (PoAC DBS) lines display more gradual curves, indicating a broader range of ARI scores. PoAC SIL starts to rise steeply around the 0.2 ARI mark, while PoAC DBS has a more pronounced curve starting from 0.1. These curves suggest that these strategies have a more variable performance, with a significant proportion of their density at lower ARI scores, indicating that they are less likely to consistently produce high-quality clustering results.

Both of these analysis reveal that the complete PoAC method is the most effective and consistent optimization strategy based on ARI values, showing high clustering quality and minimal variability. PoAC SIL shows moderate performance with significant variability, while PoAC DBS exhibits greater variability and less consistent performance. PoAC CVI shows relatively consistent performance but is not as reliable or efficient as PoAC. These findings highlight the importance of selecting the appropriate optimization strategies in relation to the specific clustering problem.

\begin{figure}[!htp]
    \centering
    \includegraphics[width=0.80\textwidth]{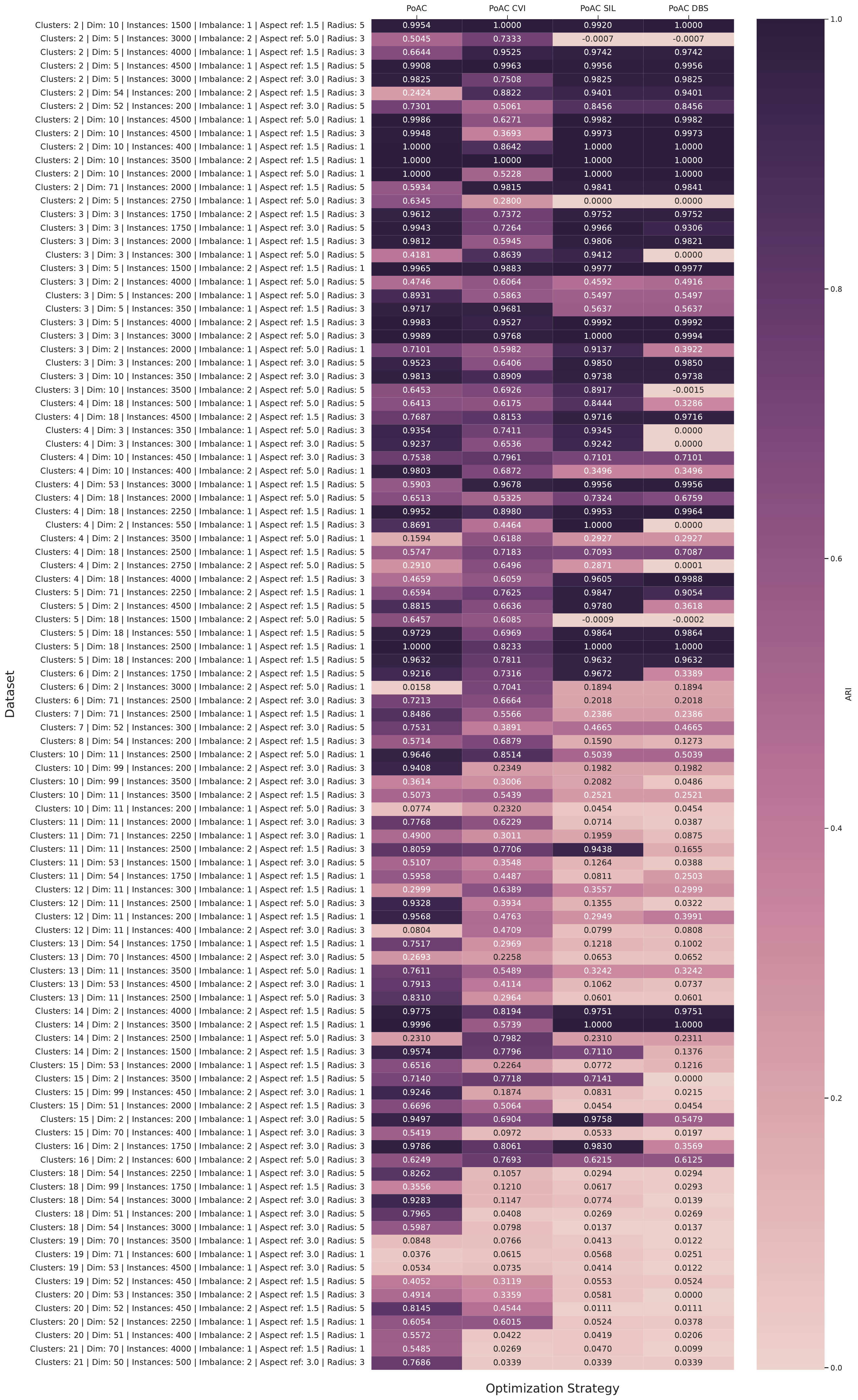}
    \caption{Optimization strategies performance per dataset.}
    \label{fig:ablatiion-heatmap}
\end{figure}

To understand the specific cases in which the optimization strategies can be at least as suited to perform as the complete PoAC method, we conducted further analysis to compare the performance of the four optimization strategies side by side, while identifying the features that define the archetype of the validation datasets, as shown in \autoref{fig:ablatiion-heatmap}. This analysis aimed to correlate the performance of the strategies with the dataset features described in \autoref{tab:validation-synth}, e.g. number of clusters, number of dimensions, and number of features. By visualizing the data in a heatmap, we were able to observe patterns and relationships that highlight how different optimization strategies perform across datasets with varying characteristics. Each cell in the heatmap represents the mean ARI for a given dataset-optimization strategy combination, with darker colors indicating higher ARI values.

For instance, the analysis revealed that strategies such as the PoAC method consistently performed well across datasets with a high number of dimensions, whereas PoAC SIL showed variable performance that was influenced more by the number of clusters. This analysis underscores the importance of considering dataset-specific features when selecting an optimization strategy for clustering, as the effectiveness of each strategy can vary significantly depending on these characteristics. This nuanced understanding enables more informed decisions when applying clustering techniques to diverse datasets, ensuring better alignment between the chosen strategy and the inherent properties of the data.

The heatmap reveals that PoAC consistently achieves high ARI values across a wide range of datasets, indicating robust performance regardless of the dataset characteristics. This strategy particularly excels in datasets with high dimensionality, a larger number of clusters, and varying levels of imbalance. PoAC maintains superior clustering quality even in datasets with higher aspect ratios (elongated clusters), and varying ratios between the maximum and minimum cluster sizes.

In contrast, the PoAC Sil strategy shows moderate performance, with variability influenced by the number of clusters and dimensions. Notably, PoAC Sil performs better in datasets with fewer clusters and lower dimensions, as well as those with lower aspect ratios and more spherical clusters. However, its performance diminishes in datasets with high aspect ratios and greater elongation. That happens because silhouette suffers from increasing dimensions since it is based on euclidean distances. So it does not handle well large dimensionalities

PoAC DBS demonstrates wide variability in performance, struggling particularly with higher dimensional datasets and those with an increased number of clusters. The inconsistent performance of PoAC DBS is evident from the lighter color cells dispersed throughout the heatmap. PoAC DBS also performs poorly in datasets with high aspect ratios and significant elongation of clusters, as well as those with high radius values.

Lastly, PoAC CVI, while generally outperforming PoAC SIL and PoAC DBS, still shows significant variability and is less reliable compared to PoAC. Its performance is more consistent in datasets with moderate to high dimensions but falls short in achieving high ARI values across all datasets. This strategy also exhibits variability in datasets with different levels of imbalance, aspect ratios, and radii, indicating a less robust performance in comparison to PoAC.


In conclusion, while sub-configurations of PoAC such as PoAC SIL and PoAC DBS are capable of achieving high values for their respective CVIs, it is important to balance them with a surrogate model. The data indicate that there are numerous instances where the highest SIL and DBS do not necessary correspond to the best partitioning results. For example, as shown in \autoref{tab:ablation_ari_sil_dbs}, PoAC SIL achieves the highest mean SIL (0.75) and PoAC DBS achieves the lowest mean DBS (0.27), yet their mean ARI values (0.53 and 0.41, respectively) are significantly lower than that of the complete PoAC method (0.70). This discrepancy underscores the need for a balanced approach that incorporates surrogate modeling to better represent the true clustering quality according to a particular training dataset. Moreover, the results for PoAC CVI, which operates without the meta-features group, further reinforce the necessity of meta-features to accurately represent the datasets and their optimal SIL and DBS values. The mean ARI for PoAC CVI (0.57) is higher than that for PoAC SIL and PoAC DBS but still lower than the complete PoAC method, demonstrating the critical role of meta-features in achieving superior clustering performance. Thus, the complete PoAC method, with its integration of surrogate modeling and meta-features, provides a more robust and effective approach to clustering optimization.

\begin{table}[!htb]
\centering
\caption{Optimization Strategies performance regarding ARI, SIL and DBS for the validation datasets.}
\begin{tabular}{lrrr}
\toprule
 \textbf{Optimization Strategy} &  \textbf{Mean ARI} &  \textbf{Mean SIL} &  \textbf{Mean DBS} \\
\midrule
\textbf{PoAC}    & \textbf{0.70} &    0.54           &    0.76 \\
        PoAC CVI &    0.57       &    0.54           &    0.68 \\
        PoAC SIL&    0.53       &    \textbf{0.75}  &    0.35 \\
        PoAC DBS&    0.41       &    0.68           &    \textbf{0.27} \\
\bottomrule
\end{tabular}
\label{tab:ablation_ari_sil_dbs}
\end{table}
\section{Limitations}
The primary limitations of this study revolve around the selection of CVIs and the assembly of datasets used during the training phase. The selection of CVIs is critical to the quality of the clustering results. The effectiveness of CVIs directly impacts the ability to accurately describe how a partitioning should be performed. Therefore, the relevance of the chosen CVIs to the specific clustering problem at hand is paramount. Higher correlation between the CVIs and the characteristics of the clustering problem (such as visualization) will yield more meaningful and effective pipeline optimization outcomes.

Furthermore, the performance of the PoAC method is influenced by the datasets used during the training phase of the surrogate model. As evidenced in the experimental results, PoAC excels when it learns clustering strategies from training datasets that have features similar to the target dataset. Consequently, the features of the training datasets must closely relate to those of the target dataset to synthesize effective clustering pipelines. This implies that careful consideration must be given to the selection and assembly of training datasets to ensure they adequately represent the diversity and characteristics of the target clustering tasks.

\section{Conclusion}
\label{Conclusion}
The PoAC framework represents a significant advancement in the field of AutoML for clustering tasks. PoAC addresses the limitations of conventional unsupervised AutoML by leveraging a comprehensive meta-knowledge base of previously encountered clustering datasets and their solutions to induce a surrogate model. This model is then utilized in the optimization process to create tailored machine learning pipelines for new, unseen datasets. This data-driven approach allows PoAC to accommodate clustering problems effectively while taking into account the specific intents of practitioners.

The main contributions of PoAC include its ability to construct a surrogate model that is finely tuned to perform specific clustering tasks with appropriate scoring mechanisms. Its algorithm-agnostic nature ensures that it can generate solutions based on the available set of algorithms without requiring additional training or data ingestion, maintaining complete independence from the optimization approach.

Experimental results demonstrated that PoAC achieves superior performance when compared to state-of-the-art frameworks across a diverse set of datasets. It also outperforms in CVIs related to visualization tasks. Additionally, PoAC has proven its capability to dynamically synthesize clustering pipelines by adjusting preprocessing steps based on the complexity of the dataset and the defined problem.

Future work on the PoAC framework will focus on expanding its applicability to tackle additional clustering challenges, specifically scalability and noise reduction. As datasets continue to grow in size and complexity, developing scalable clustering solutions that maintain performance and efficiency is crucial. We aim to explore advanced strategies that can enhance PoAC's capability to process large-scale datasets without compromising clustering quality. Additionally, addressing noise reduction in clustering will be a key area of investigation, as the presence of noise can significantly impact the effectiveness of clustering algorithms. By integrating techniques for robust noise handling into the PoAC framework, we hope to improve the framework’s resilience and adaptability across various real-world scenarios, ultimately providing users with more reliable and context-sensitive clustering solutions.








\bigskip
\begin{flushleft}%

\bigskip\noindent
\textbf{Author contributions:} \textit{Conceptualization}: Matheus Camilo da Silva, Sylvio Barbon Junior;
\textit{Formal analysis and investigation}: Matheus Camilo da Silva, Gabriel Marques Tavares, Eric Medvet, Sylvio Barbon Junior;
\textit{Writing—original draft preparation}: Matheus Camilo da Silva;
\textit{Writing—review and editing}: Matheus Camilo da Silva, Gabriel Marques Tavares, Eric Medvet, Sylvio Barbon Junior;
\textit{Supervision}: Sylvio Barbon Junior.

\bigskip\noindent
\textbf{Funding:} The author Matheus Camilo da Silva receives funding from the Ministero dell'Università e della Ricerca (Italian Ministry of University and Research) in the form of a PhD scholarship.

\end{flushleft}



\section*{Declarations}
\begin{itemize}
    \item \textbf{Confict of interest} The authors have no Confict of interest to declare that are relevant to the content of this article.
    \item  \textbf{Ethical approval} Not applicable.
    \item \textbf{Consent to participate} Not applicable.
    \item \textbf{Consent for publication} Not applicable.
\end{itemize}


\noindent

\bigskip
\begin{appendices}






\end{appendices}


\bibliography{sn-bibliography}


\begin{thebibliography}{58}
\ifx \bisbn   \undefined \def \bisbn  #1{ISBN #1}\fi
\ifx \binits  \undefined \def \binits#1{#1}\fi
\ifx \bauthor  \undefined \def \bauthor#1{#1}\fi
\ifx \batitle  \undefined \def \batitle#1{#1}\fi
\ifx \bjtitle  \undefined \def \bjtitle#1{#1}\fi
\ifx \bvolume  \undefined \def \bvolume#1{\textbf{#1}}\fi
\ifx \byear  \undefined \def \byear#1{#1}\fi
\ifx \bissue  \undefined \def \bissue#1{#1}\fi
\ifx \bfpage  \undefined \def \bfpage#1{#1}\fi
\ifx \blpage  \undefined \def \blpage #1{#1}\fi
\ifx \burl  \undefined \def \burl#1{\textsf{#1}}\fi
\ifx \doiurl  \undefined \def \doiurl#1{\url{https://doi.org/#1}}\fi
\ifx \betal  \undefined \def \betal{\textit{et al.}}\fi
\ifx \binstitute  \undefined \def \binstitute#1{#1}\fi
\ifx \binstitutionaled  \undefined \def \binstitutionaled#1{#1}\fi
\ifx \bctitle  \undefined \def \bctitle#1{#1}\fi
\ifx \beditor  \undefined \def \beditor#1{#1}\fi
\ifx \bpublisher  \undefined \def \bpublisher#1{#1}\fi
\ifx \bbtitle  \undefined \def \bbtitle#1{#1}\fi
\ifx \bedition  \undefined \def \bedition#1{#1}\fi
\ifx \bseriesno  \undefined \def \bseriesno#1{#1}\fi
\ifx \blocation  \undefined \def \blocation#1{#1}\fi
\ifx \bsertitle  \undefined \def \bsertitle#1{#1}\fi
\ifx \bsnm \undefined \def \bsnm#1{#1}\fi
\ifx \bsuffix \undefined \def \bsuffix#1{#1}\fi
\ifx \bparticle \undefined \def \bparticle#1{#1}\fi
\ifx \barticle \undefined \def \barticle#1{#1}\fi
\bibcommenthead
\ifx \bconfdate \undefined \def \bconfdate #1{#1}\fi
\ifx \botherref \undefined \def \botherref #1{#1}\fi
\ifx \url \undefined \def \url#1{\textsf{#1}}\fi
\ifx \bchapter \undefined \def \bchapter#1{#1}\fi
\ifx \bbook \undefined \def \bbook#1{#1}\fi
\ifx \bcomment \undefined \def \bcomment#1{#1}\fi
\ifx \oauthor \undefined \def \oauthor#1{#1}\fi
\ifx \citeauthoryear \undefined \def \citeauthoryear#1{#1}\fi
\ifx \endbibitem  \undefined \def \endbibitem {}\fi
\ifx \bconflocation  \undefined \def \bconflocation#1{#1}\fi
\ifx \arxivurl  \undefined \def \arxivurl#1{\textsf{#1}}\fi
\csname PreBibitemsHook\endcsname

\bibitem[\protect\citeauthoryear{Al-Jabery et~al.}{2019}]{al2019computational}
\begin{bbook}
\bauthor{\bsnm{Al-Jabery}, \binits{K.}},
\bauthor{\bsnm{Obafemi-Ajayi}, \binits{T.}},
\bauthor{\bsnm{Olbricht}, \binits{G.}},
\bauthor{\bsnm{Wunsch}, \binits{D.}}:
\bbtitle{Computational Learning Approaches to Data Analytics in Biomedical Applications}.
\bpublisher{Academic Press}, \blocation{???}
(\byear{2019})
\end{bbook}
\endbibitem

\bibitem[\protect\citeauthoryear{Assent}{2012}]{assent2012clustering}
\begin{barticle}
\bauthor{\bsnm{Assent}, \binits{I.}}:
\batitle{Clustering high dimensional data}.
\bjtitle{Wiley Interdisciplinary Reviews: Data Mining and Knowledge Discovery}
\bvolume{2}(\bissue{4}),
\bfpage{340}--\blpage{350}
(\byear{2012})
\end{barticle}
\endbibitem

\bibitem[\protect\citeauthoryear{Bagirov et~al.}{2023}]{BAGIROV2023109144}
\begin{barticle}
\bauthor{\bsnm{Bagirov}, \binits{A.M.}},
\bauthor{\bsnm{Aliguliyev}, \binits{R.M.}},
\bauthor{\bsnm{Sultanova}, \binits{N.}}:
\batitle{Finding compact and well-separated clusters: Clustering using silhouette coefficients}.
\bjtitle{Pattern Recognition}
\bvolume{135},
\bfpage{109144}
(\byear{2023})
\doiurl{10.1016/j.patcog.2022.109144}
\end{barticle}
\endbibitem

\bibitem[\protect\citeauthoryear{Bonner}{1964}]{bonner1964some}
\begin{barticle}
\bauthor{\bsnm{Bonner}, \binits{R.E.}}:
\batitle{On some clustering techniques}.
\bjtitle{IBM journal of research and development}
\bvolume{8}(\bissue{1}),
\bfpage{22}--\blpage{32}
(\byear{1964})
\end{barticle}
\endbibitem

\bibitem[\protect\citeauthoryear{Breiman}{2001}]{breiman2001random}
\begin{barticle}
\bauthor{\bsnm{Breiman}, \binits{L.}}:
\batitle{Random forests}.
\bjtitle{Machine Learning}
\bvolume{45}(\bissue{1}),
\bfpage{5}--\blpage{32}
(\byear{2001})
\doiurl{10.1023/A:1010933404324}
\end{barticle}
\endbibitem

\bibitem[\protect\citeauthoryear{Bahri et~al.}{2022}]{bahri2022automl}
\begin{barticle}
\bauthor{\bsnm{Bahri}, \binits{M.}},
\bauthor{\bsnm{Salutari}, \binits{F.}},
\bauthor{\bsnm{Putina}, \binits{A.}},
\bauthor{\bsnm{Sozio}, \binits{M.}}:
\batitle{Automl: state of the art with a focus on anomaly detection, challenges, and research directions}.
\bjtitle{International Journal of Data Science and Analytics}
\bvolume{14}(\bissue{2}),
\bfpage{113}--\blpage{126}
(\byear{2022})
\end{barticle}
\endbibitem

\bibitem[\protect\citeauthoryear{Brazdil et~al.}{2022}]{brazdil2022meta}
\begin{bbook}
\bauthor{\bsnm{Brazdil}, \binits{P.}},
\bauthor{\bsnm{Rijn}, \binits{J.N.}},
\bauthor{\bsnm{Soares}, \binits{C.}},
\bauthor{\bsnm{Vanschoren}, \binits{J.}}:
\bbtitle{Metalearning: Applications to Automated Machine Learning and Data Mining}.
\bpublisher{Springer}, \blocation{???}
(\byear{2022}).
\doiurl{10.1007/978-3-030-67024-5} .
\burl{https://doi.org/10.1007/978-3-030-67024-5}
\end{bbook}
\endbibitem

\bibitem[\protect\citeauthoryear{Chatzimparmpas et~al.}{2020}]{chatzimparmpas2020survey}
\begin{barticle}
\bauthor{\bsnm{Chatzimparmpas}, \binits{A.}},
\bauthor{\bsnm{Martins}, \binits{R.M.}},
\bauthor{\bsnm{Jusufi}, \binits{I.}},
\bauthor{\bsnm{Kerren}, \binits{A.}}:
\batitle{A survey of surveys on the use of visualization for interpreting machine learning models}.
\bjtitle{Information Visualization}
\bvolume{19}(\bissue{3}),
\bfpage{207}--\blpage{233}
(\byear{2020})
\end{barticle}
\endbibitem

\bibitem[\protect\citeauthoryear{Cohen-Shapira and Rokach}{2021}]{cohenshapira2021automatic}
\begin{barticle}
\bauthor{\bsnm{Cohen-Shapira}, \binits{N.}},
\bauthor{\bsnm{Rokach}, \binits{L.}}:
\batitle{Automatic selection of clustering algorithms using supervised graph embedding}.
\bjtitle{Information Sciences}
\bvolume{577},
\bfpage{824}--\blpage{851}
(\byear{2021})
\doiurl{10.1016/j.ins.2021.08.028}
\end{barticle}
\endbibitem

\bibitem[\protect\citeauthoryear{Davies and Bouldin}{1979}]{davies1979cluster}
\begin{botherref}
\oauthor{\bsnm{Davies}, \binits{D.L.}},
\oauthor{\bsnm{Bouldin}, \binits{D.W.}}:
A cluster separation measure.
IEEE transactions on pattern analysis and machine intelligence
(2),
224--227
(1979)
\end{botherref}
\endbibitem

\bibitem[\protect\citeauthoryear{da~Silva et~al.}{2024}]{silva2024benchmarking}
\begin{bchapter}
\bauthor{\bsnm{Silva}, \binits{M.C.}},
\bauthor{\bsnm{Licari}, \binits{B.}},
\bauthor{\bsnm{Tavares}, \binits{G.M.}},
\bauthor{\bsnm{Junior}, \binits{S.B.}}:
\bctitle{Benchmarking auto{ML} clustering frameworks}.
In: \bbtitle{AutoML Conference 2024 (ABCD Track)}
(\byear{2024}).
\burl{https://openreview.net/forum?id=RzUKJnph1g}
\end{bchapter}
\endbibitem

\bibitem[\protect\citeauthoryear{de~Souto et~al.}{2008}]{souto2008ranking}
\begin{bchapter}
\bauthor{\bsnm{Souto}, \binits{M.C.P.}},
\bauthor{\bsnm{Prudencio}, \binits{R.B.C.}},
\bauthor{\bsnm{Soares}, \binits{R.G.F.}},
\bauthor{\bsnm{Araujo}, \binits{D.S.A.}},
\bauthor{\bsnm{Costa}, \binits{I.G.}},
\bauthor{\bsnm{Ludermir}, \binits{T.B.}},
\bauthor{\bsnm{Schliep}, \binits{A.}}:
\bctitle{Ranking and selecting clustering algorithms using a meta-learning approach}.
In: \bbtitle{2008 IEEE International Joint Conference on Neural Networks (IEEE World Congress on Computational Intelligence)},
pp. \bfpage{3729}--\blpage{3735}
(\byear{2008}).
\doiurl{10.1109/IJCNN.2008.4634333}
\end{bchapter}
\endbibitem

\bibitem[\protect\citeauthoryear{Ezugwu et~al.}{2022}]{ezugwu2022comprehensive}
\begin{barticle}
\bauthor{\bsnm{Ezugwu}, \binits{A.E.}},
\bauthor{\bsnm{Ikotun}, \binits{A.M.}},
\bauthor{\bsnm{Oyelade}, \binits{O.O.}},
\bauthor{\bsnm{Abualigah}, \binits{L.}},
\bauthor{\bsnm{Agushaka}, \binits{J.O.}},
\bauthor{\bsnm{Eke}, \binits{C.I.}},
\bauthor{\bsnm{Akinyelu}, \binits{A.A.}}:
\batitle{A comprehensive survey of clustering algorithms: State-of-the-art machine learning applications, taxonomy, challenges, and future research prospects}.
\bjtitle{Engineering Applications of Artificial Intelligence}
\bvolume{110},
\bfpage{104743}
(\byear{2022})
\end{barticle}
\endbibitem

\bibitem[\protect\citeauthoryear{ElShawi et~al.}{2021}]{elshawi2021csmartml}
\begin{bchapter}
\bauthor{\bsnm{ElShawi}, \binits{R.}},
\bauthor{\bsnm{Lekunze}, \binits{H.}},
\bauthor{\bsnm{Sakr}, \binits{S.}}:
\bctitle{csmartml: A meta learning-based framework for automated selection and hyperparameter tuning for clustering}.
In: \bbtitle{2021 IEEE International Conference on Big Data (Big Data)},
pp. \bfpage{1119}--\blpage{1126}
(\byear{2021}).
\doiurl{10.1109/BigData52589.2021.9671542}
\end{bchapter}
\endbibitem

\bibitem[\protect\citeauthoryear{ElShawi and Sakr}{2022}]{elshawi2022tpe}
\begin{bchapter}
\bauthor{\bsnm{ElShawi}, \binits{R.}},
\bauthor{\bsnm{Sakr}, \binits{S.}}:
\bctitle{Tpe-autoclust: A tree-based pipline ensemble framework for automated clustering}.
In: \bbtitle{2022 IEEE International Conference on Data Mining Workshops (ICDMW)},
pp. \bfpage{1144}--\blpage{1153}
(\byear{2022}).
\doiurl{10.1109/ICDMW58026.2022.00149}
\end{bchapter}
\endbibitem

\bibitem[\protect\citeauthoryear{Ferrari and {de Castro}}{2015}]{ferrari2015clustering}
\begin{barticle}
\bauthor{\bsnm{Ferrari}, \binits{D.G.}},
\bauthor{\bsnm{{de Castro}}, \binits{L.N.}}:
\batitle{Clustering algorithm selection by meta-learning systems: A new distance-based problem characterization and ranking combination methods}.
\bjtitle{Information Sciences}
\bvolume{301},
\bfpage{181}--\blpage{194}
(\byear{2015})
\doiurl{10.1016/j.ins.2014.12.044}
\end{barticle}
\endbibitem

\bibitem[\protect\citeauthoryear{Ferrari and de~Castro}{2012}]{ferrari2012clustering}
\begin{bchapter}
\bauthor{\bsnm{Ferrari}, \binits{D.G.}},
\bauthor{\bsnm{Castro}, \binits{L.N.}}:
\bctitle{Clustering algorithm recommendation: A meta-learning approach}.
In: \beditor{\bsnm{Panigrahi}, \binits{B.K.}},
\beditor{\bsnm{Das}, \binits{S.}},
\beditor{\bsnm{Suganthan}, \binits{P.N.}},
\beditor{\bsnm{Nanda}, \binits{P.K.}} (eds.)
\bbtitle{Swarm, Evolutionary, and Memetic Computing},
pp. \bfpage{143}--\blpage{150}.
\bpublisher{Springer},
\blocation{Berlin, Heidelberg}
(\byear{2012})
\end{bchapter}
\endbibitem

\bibitem[\protect\citeauthoryear{Fernandes et~al.}{2021}]{fernandes2021towards}
\begin{botherref}
\oauthor{\bsnm{Fernandes}, \binits{L.H.d.S.}},
\oauthor{\bsnm{Lorena}, \binits{A.C.}},
\oauthor{\bsnm{Smith-Miles}, \binits{K.}}:
Towards understanding clustering problems and algorithms: An instance space analysis.
Algorithms
\textbf{14}(3)
(2021)
\doiurl{10.3390/a14030095}
\end{botherref}
\endbibitem

\bibitem[\protect\citeauthoryear{Gabbay et~al.}{2021}]{gabbay2021isolation}
\begin{barticle}
\bauthor{\bsnm{Gabbay}, \binits{I.}},
\bauthor{\bsnm{Shapira}, \binits{B.}},
\bauthor{\bsnm{Rokach}, \binits{L.}}:
\batitle{Isolation forests and landmarking-based representations for clustering algorithm recommendation using meta-learning}.
\bjtitle{Information Sciences}
\bvolume{574},
\bfpage{473}--\blpage{489}
(\byear{2021})
\doiurl{10.1016/j.ins.2021.06.033}
\end{barticle}
\endbibitem

\bibitem[\protect\citeauthoryear{Hubert and Arabie}{1985}]{hubert1985comparing}
\begin{barticle}
\bauthor{\bsnm{Hubert}, \binits{L.}},
\bauthor{\bsnm{Arabie}, \binits{P.}}:
\batitle{Comparing partitions}.
\bjtitle{Journal of classification}
\bvolume{2},
\bfpage{193}--\blpage{218}
(\byear{1985})
\end{barticle}
\endbibitem

\bibitem[\protect\citeauthoryear{Hennig}{2015}]{hennig2015what}
\begin{barticle}
\bauthor{\bsnm{Hennig}, \binits{C.}}:
\batitle{What are the true clusters?}
\bjtitle{Pattern Recognition Letters}
\bvolume{64},
\bfpage{53}--\blpage{62}
(\byear{2015})
\doiurl{10.1016/j.patrec.2015.04.009} .
\bcomment{Philosophical Aspects of Pattern Recognition}
\end{barticle}
\endbibitem

\bibitem[\protect\citeauthoryear{Hutter et~al.}{2019}]{hutter2019automated}
\begin{bbook}
\bauthor{\bsnm{Hutter}, \binits{F.}},
\bauthor{\bsnm{Kotthoff}, \binits{L.}},
\bauthor{\bsnm{Vanschoren}, \binits{J.}}:
\bbtitle{Automated Machine Learning: Methods, Systems, Challenges}.
\bpublisher{Springer}, \blocation{???}
(\byear{2019})
\end{bbook}
\endbibitem

\bibitem[\protect\citeauthoryear{Hastie et~al.}{2009}]{hastie2009random}
\begin{botherref}
\oauthor{\bsnm{Hastie}, \binits{T.}},
\oauthor{\bsnm{Tibshirani}, \binits{R.}},
\oauthor{\bsnm{Friedman}, \binits{J.}},
\oauthor{\bsnm{Hastie}, \binits{T.}},
\oauthor{\bsnm{Tibshirani}, \binits{R.}},
\oauthor{\bsnm{Friedman}, \binits{J.}}:
Random forests.
The elements of statistical learning: Data mining, inference, and prediction,
587--604
(2009)
\end{botherref}
\endbibitem

\bibitem[\protect\citeauthoryear{He et~al.}{2021}]{he2021automl}
\begin{barticle}
\bauthor{\bsnm{He}, \binits{X.}},
\bauthor{\bsnm{Zhao}, \binits{K.}},
\bauthor{\bsnm{Chu}, \binits{X.}}:
\batitle{Automl: A survey of the state-of-the-art}.
\bjtitle{Knowledge-Based Systems}
\bvolume{212},
\bfpage{106622}
(\byear{2021})
\end{barticle}
\endbibitem

\bibitem[\protect\citeauthoryear{Ilter and Guvenir}{1998}]{misc_dermatology_33}
\begin{botherref}
\oauthor{\bsnm{Ilter}, \binits{N.}},
\oauthor{\bsnm{Guvenir}, \binits{H.}}:
{Dermatology}.
UCI Machine Learning Repository.
{DOI}: https://doi.org/10.24432/C5FK5P
(1998)
\end{botherref}
\endbibitem

\bibitem[\protect\citeauthoryear{Jolliffe and Cadima}{2016}]{jolliffe2016principal}
\begin{barticle}
\bauthor{\bsnm{Jolliffe}, \binits{I.T.}},
\bauthor{\bsnm{Cadima}, \binits{J.}}:
\batitle{Principal component analysis: a review and recent developments}.
\bjtitle{Philosophical transactions of the royal society A: Mathematical, Physical and Engineering Sciences}
\bvolume{374}(\bissue{2065}),
\bfpage{20150202}
(\byear{2016})
\end{barticle}
\endbibitem

\bibitem[\protect\citeauthoryear{Kryszczuk and Hurley}{2010}]{Kryszczuk2010EstimationOT}
\begin{bchapter}
\bauthor{\bsnm{Kryszczuk}, \binits{K.}},
\bauthor{\bsnm{Hurley}, \binits{P.}}:
\bctitle{Estimation of the number of clusters using multiple clustering validity indices}.
In: \bbtitle{International Workshop on Multiple Classifier Systems}
(\byear{2010}).
\burl{https://api.semanticscholar.org/CorpusID:17705188}
\end{bchapter}
\endbibitem

\bibitem[\protect\citeauthoryear{Lorena et~al.}{2019}]{lorena2019how}
\begin{botherref}
\oauthor{\bsnm{Lorena}, \binits{A.C.}},
\oauthor{\bsnm{Garcia}, \binits{L.P.F.}},
\oauthor{\bsnm{Lehmann}, \binits{J.}},
\oauthor{\bsnm{Souto}, \binits{M.C.P.}},
\oauthor{\bsnm{Ho}, \binits{T.K.}}:
How complex is your classification problem? a survey on measuring classification complexity.
ACM Comput. Surv.
\textbf{52}(5)
(2019)
\doiurl{10.1145/3347711}
\end{botherref}
\endbibitem

\bibitem[\protect\citeauthoryear{Liu et~al.}{2021}]{liu2021autocluster}
\begin{bchapter}
\bauthor{\bsnm{Liu}, \binits{Y.}},
\bauthor{\bsnm{Li}, \binits{S.}},
\bauthor{\bsnm{Tian}, \binits{W.}}:
\bctitle{Autocluster: Meta-learning based ensemble method for automated unsupervised clustering}.
In: \beditor{\bsnm{Karlapalem}, \binits{K.}},
\beditor{\bsnm{Cheng}, \binits{H.}},
\beditor{\bsnm{Ramakrishnan}, \binits{N.}},
\beditor{\bsnm{Agrawal}, \binits{R.K.}},
\beditor{\bsnm{Reddy}, \binits{P.K.}},
\beditor{\bsnm{Srivastava}, \binits{J.}},
\beditor{\bsnm{Chakraborty}, \binits{T.}} (eds.)
\bbtitle{Advances in Knowledge Discovery and Data Mining},
pp. \bfpage{246}--\blpage{258}.
\bpublisher{Springer},
\blocation{Cham}
(\byear{2021})
\end{bchapter}
\endbibitem

\bibitem[\protect\citeauthoryear{Lopes et~al.}{2019}]{lopes2019improving}
\begin{botherref}
\oauthor{\bsnm{Lopes}, \binits{R.G.}},
\oauthor{\bsnm{Yin}, \binits{D.}},
\oauthor{\bsnm{Poole}, \binits{B.}},
\oauthor{\bsnm{Gilmer}, \binits{J.}},
\oauthor{\bsnm{Cubuk}, \binits{E.D.}}:
Improving robustness without sacrificing accuracy with patch gaussian augmentation.
arXiv preprint arXiv:1906.02611
(2019)
\end{botherref}
\endbibitem

\bibitem[\protect\citeauthoryear{Maulik and Bandyopadhyay}{2002}]{Maulik2002PerformanceEO}
\begin{barticle}
\bauthor{\bsnm{Maulik}, \binits{U.}},
\bauthor{\bsnm{Bandyopadhyay}, \binits{S.}}:
\batitle{Performance evaluation of some clustering algorithms and validity indices}.
\bjtitle{IEEE Trans. Pattern Anal. Mach. Intell.}
\bvolume{24},
\bfpage{1650}--\blpage{1654}
(\byear{2002})
\end{barticle}
\endbibitem

\bibitem[\protect\citeauthoryear{Marutho et~al.}{2018}]{Marutho2018TheDO}
\begin{botherref}
\oauthor{\bsnm{Marutho}, \binits{D.}},
\oauthor{\bsnm{Handaka}, \binits{S.H.}},
\oauthor{\bsnm{Wijaya}, \binits{E.}},
\oauthor{\bsnm{Muljono}}:
The determination of cluster number at k-mean using elbow method and purity evaluation on headline news.
2018 International Seminar on Application for Technology of Information and Communication,
533--538
(2018)
\end{botherref}
\endbibitem

\bibitem[\protect\citeauthoryear{Markelle~Kelly}{2017}]{uci@2017}
\begin{botherref}
\oauthor{\bsnm{Markelle~Kelly}, \binits{K.N.} \bsuffix{Rachel~Longjohn}}:
UCI Machine Learning Repository
(2017).
\url{http://archive.ics.uci.edu/ml}
\end{botherref}
\endbibitem

\bibitem[\protect\citeauthoryear{Mishra et~al.}{2022}]{mishra2022evaluative}
\begin{barticle}
\bauthor{\bsnm{Mishra}, \binits{S.}},
\bauthor{\bsnm{Monath}, \binits{N.}},
\bauthor{\bsnm{Boratko}, \binits{M.}},
\bauthor{\bsnm{Kobren}, \binits{A.}},
\bauthor{\bsnm{McCallum}, \binits{A.}}:
\batitle{An evaluative measure of clustering methods incorporating hyperparameter sensitivity}.
\bjtitle{Proceedings of the AAAI Conference on Artificial Intelligence}
\bvolume{36}(\bissue{7}),
\bfpage{7788}--\blpage{7796}
(\byear{2022})
\doiurl{10.1609/aaai.v36i7.20747}
\end{barticle}
\endbibitem

\bibitem[\protect\citeauthoryear{Nemenyi}{1963}]{nemenyi1963distribution}
\begin{bbook}
\bauthor{\bsnm{Nemenyi}, \binits{P.B.}}:
\bbtitle{Distribution-free Multiple Comparisons.}
\bpublisher{Princeton University}, \blocation{???}
(\byear{1963})
\end{bbook}
\endbibitem

\bibitem[\protect\citeauthoryear{Nascimento et~al.}{2009}]{nascimento2009mining}
\begin{bchapter}
\bauthor{\bsnm{Nascimento}, \binits{A.C.A.}},
\bauthor{\bsnm{Prud{\^e}ncio}, \binits{R.B.C.}},
\bauthor{\bsnm{Souto}, \binits{M.C.P.}},
\bauthor{\bsnm{Costa}, \binits{I.G.}}:
\bctitle{Mining rules for the automatic selection process of clustering methods applied to cancer gene expression data}.
In: \beditor{\bsnm{Alippi}, \binits{C.}},
\beditor{\bsnm{Polycarpou}, \binits{M.}},
\beditor{\bsnm{Panayiotou}, \binits{C.}},
\beditor{\bsnm{Ellinas}, \binits{G.}} (eds.)
\bbtitle{Artificial Neural Networks -- ICANN 2009},
pp. \bfpage{20}--\blpage{29}.
\bpublisher{Springer},
\blocation{Berlin, Heidelberg}
(\byear{2009})
\end{bchapter}
\endbibitem

\bibitem[\protect\citeauthoryear{Olson and Moore}{2016}]{olson2016tpot}
\begin{bchapter}
\bauthor{\bsnm{Olson}, \binits{R.S.}},
\bauthor{\bsnm{Moore}, \binits{J.H.}}:
\bctitle{Tpot: A tree-based pipeline optimization tool for automating machine learning}.
In: \bbtitle{Workshop on Automatic Machine Learning},
pp. \bfpage{66}--\blpage{74}
(\byear{2016}).
\bcomment{PMLR}
\end{bchapter}
\endbibitem

\bibitem[\protect\citeauthoryear{Olson et~al.}{2016}]{Olson2016AutomatingBD}
\begin{bchapter}
\bauthor{\bsnm{Olson}, \binits{R.S.}},
\bauthor{\bsnm{Urbanowicz}, \binits{R.J.}},
\bauthor{\bsnm{Andrews}, \binits{P.C.}},
\bauthor{\bsnm{Lavender}, \binits{N.A.}},
\bauthor{\bsnm{Kidd}, \binits{L.C.}},
\bauthor{\bsnm{Moore}, \binits{J.H.}}:
\bctitle{Automating biomedical data science through tree-based pipeline optimization}.
In: \bbtitle{EvoApplications}
(\byear{2016}).
\burl{https://api.semanticscholar.org/CorpusID:9709316}
\end{bchapter}
\endbibitem

\bibitem[\protect\citeauthoryear{Pimentel and {de Carvalho}}{2019}]{pimentel2019new}
\begin{barticle}
\bauthor{\bsnm{Pimentel}, \binits{B.A.}},
\bauthor{\bsnm{{de Carvalho}}, \binits{A.C.P.L.F.}}:
\batitle{A new data characterization for selecting clustering algorithms using meta-learning}.
\bjtitle{Information Sciences}
\bvolume{477},
\bfpage{203}--\blpage{219}
(\byear{2019})
\doiurl{10.1016/j.ins.2018.10.043}
\end{barticle}
\endbibitem

\bibitem[\protect\citeauthoryear{Pimentel and de~Carvalho}{2018}]{pimentel2018statistical}
\begin{bchapter}
\bauthor{\bsnm{Pimentel}, \binits{B.A.}},
\bauthor{\bsnm{Carvalho}, \binits{A.C.P.L.F.}}:
\bctitle{Statistical versus distance-based meta-features for clustering algorithm recommendation using meta-learning}.
In: \bbtitle{2018 International Joint Conference on Neural Networks (IJCNN)},
pp. \bfpage{1}--\blpage{8}
(\byear{2018}).
\doiurl{10.1109/IJCNN.2018.8489182}
\end{bchapter}
\endbibitem

\bibitem[\protect\citeauthoryear{Pimentel and de~Carvalho}{2019}]{pimentel2019unsupervised}
\begin{bchapter}
\bauthor{\bsnm{Pimentel}, \binits{B.A.}},
\bauthor{\bsnm{Carvalho}, \binits{A.C.P.L.F.}}:
\bctitle{Unsupervised meta-learning for clustering algorithm recommendation}.
In: \bbtitle{2019 International Joint Conference on Neural Networks (IJCNN)},
pp. \bfpage{1}--\blpage{8}
(\byear{2019}).
\doiurl{10.1109/IJCNN.2019.8851989}
\end{bchapter}
\endbibitem

\bibitem[\protect\citeauthoryear{Poulakis et~al.}{2020}]{poulakis2020autoclust}
\begin{bchapter}
\bauthor{\bsnm{Poulakis}, \binits{Y.}},
\bauthor{\bsnm{Doulkeridis}, \binits{C.}},
\bauthor{\bsnm{Kyriazis}, \binits{D.}}:
\bctitle{Autoclust: A framework for automated clustering based on cluster validity indices}.
In: \bbtitle{2020 IEEE International Conference on Data Mining (ICDM)},
pp. \bfpage{1220}--\blpage{1225}
(\byear{2020}).
\bcomment{IEEE}
\end{bchapter}
\endbibitem

\bibitem[\protect\citeauthoryear{Rice}{1976}]{rice1976algorithm}
\begin{bchapter}
\bauthor{\bsnm{Rice}, \binits{J.R.}}:
\bctitle{The algorithm selection problem}.
In: \bbtitle{Advances in Computers}
vol. \bseriesno{15},
pp. \bfpage{65}--\blpage{118}.
\bpublisher{Elsevier}, \blocation{???}
(\byear{1976})
\end{bchapter}
\endbibitem

\bibitem[\protect\citeauthoryear{Rokach and Maimon}{2005}]{rokach2005clustering}
\begin{botherref}
\oauthor{\bsnm{Rokach}, \binits{L.}},
\oauthor{\bsnm{Maimon}, \binits{O.}}:
Clustering methods.
Data mining and knowledge discovery handbook,
321--352
(2005)
\end{botherref}
\endbibitem

\bibitem[\protect\citeauthoryear{Rousseeuw}{1987}]{rousseeuw1987silhouettes}
\begin{barticle}
\bauthor{\bsnm{Rousseeuw}, \binits{P.J.}}:
\batitle{Silhouettes: a graphical aid to the interpretation and validation of cluster analysis}.
\bjtitle{Journal of computational and applied mathematics}
\bvolume{20},
\bfpage{53}--\blpage{65}
(\byear{1987})
\end{barticle}
\endbibitem

\bibitem[\protect\citeauthoryear{Strehl and Ghosh}{2003}]{strehl2003relationship}
\begin{barticle}
\bauthor{\bsnm{Strehl}, \binits{A.}},
\bauthor{\bsnm{Ghosh}, \binits{J.}}:
\batitle{Relationship-based clustering and visualization for high-dimensional data mining}.
\bjtitle{INFORMS Journal on Computing}
\bvolume{15}(\bissue{2}),
\bfpage{208}--\blpage{230}
(\byear{2003})
\end{barticle}
\endbibitem

\bibitem[\protect\citeauthoryear{Soares et~al.}{2009}]{soares2009analysis}
\begin{bchapter}
\bauthor{\bsnm{Soares}, \binits{R.G.F.}},
\bauthor{\bsnm{Ludermir}, \binits{T.B.}},
\bauthor{\bsnm{De~Carvalho}, \binits{F.A.T.}}:
\bctitle{An analysis of meta-learning techniques for ranking clustering algorithms applied to artificial data}.
In: \beditor{\bsnm{Alippi}, \binits{C.}},
\beditor{\bsnm{Polycarpou}, \binits{M.}},
\beditor{\bsnm{Panayiotou}, \binits{C.}},
\beditor{\bsnm{Ellinas}, \binits{G.}} (eds.)
\bbtitle{Artificial Neural Networks -- ICANN 2009},
pp. \bfpage{131}--\blpage{140}.
\bpublisher{Springer},
\blocation{Berlin, Heidelberg}
(\byear{2009})
\end{bchapter}
\endbibitem

\bibitem[\protect\citeauthoryear{Shahapure and Nicholas}{2020}]{Shahapure2020ClusterQA}
\begin{botherref}
\oauthor{\bsnm{Shahapure}, \binits{K.R.}},
\oauthor{\bsnm{Nicholas}, \binits{C.K.}}:
Cluster quality analysis using silhouette score.
2020 IEEE 7th International Conference on Data Science and Advanced Analytics (DSAA),
747--748
(2020)
\end{botherref}
\endbibitem

\bibitem[\protect\citeauthoryear{Tschechlov et~al.}{2021}]{tschechlov2021automl4clust}
\begin{botherref}
\oauthor{\bsnm{Tschechlov}, \binits{D.}},
\oauthor{\bsnm{Fritz}, \binits{M.}},
\oauthor{\bsnm{Schwarz}, \binits{H.}}:
AutoML4Clust: Efficient AutoML for Clustering Analyses.
OpenProceedings.org
(2021).
\doiurl{10.5441/002/EDBT.2021.32} .
\url{https://openproceedings.org/2021/conf/edbt/p87.pdf}
\end{botherref}
\endbibitem

\bibitem[\protect\citeauthoryear{Thomas et~al.}{2013}]{Thomas2013NewVO}
\begin{botherref}
\oauthor{\bsnm{Thomas}, \binits{J.C.R.}},
\oauthor{\bsnm{Pe{\~n}as}, \binits{M.S.}},
\oauthor{\bsnm{Mora}, \binits{M.}}:
New version of davies-bouldin index for clustering validation based on cylindrical distance.
2013 32nd International Conference of the Chilean Computer Science Society (SCCC),
49--53
(2013)
\end{botherref}
\endbibitem

\bibitem[\protect\citeauthoryear{Treder-Tschechlov et~al.}{2023}]{treder2023ml2dac}
\begin{barticle}
\bauthor{\bsnm{Treder-Tschechlov}, \binits{D.}},
\bauthor{\bsnm{Fritz}, \binits{M.}},
\bauthor{\bsnm{Schwarz}, \binits{H.}},
\bauthor{\bsnm{Mitschang}, \binits{B.}}:
\batitle{Ml2dac: Meta-learning to democratize automl for clustering analysis}.
\bjtitle{Proceedings of the ACM on Management of Data}
\bvolume{1}(\bissue{2}),
\bfpage{1}--\blpage{26}
(\byear{2023})
\end{barticle}
\endbibitem

\bibitem[\protect\citeauthoryear{Truong et~al.}{2019}]{truong2019towards}
\begin{bchapter}
\bauthor{\bsnm{Truong}, \binits{A.}},
\bauthor{\bsnm{Walters}, \binits{A.}},
\bauthor{\bsnm{Goodsitt}, \binits{J.}},
\bauthor{\bsnm{Hines}, \binits{K.}},
\bauthor{\bsnm{Bruss}, \binits{C.B.}},
\bauthor{\bsnm{Farivar}, \binits{R.}}:
\bctitle{Towards automated machine learning: Evaluation and comparison of automl approaches and tools}.
In: \bbtitle{2019 IEEE 31st International Conference on Tools with Artificial Intelligence (ICTAI)},
pp. \bfpage{1471}--\blpage{1479}
(\byear{2019}).
\bcomment{IEEE}
\end{bchapter}
\endbibitem

\bibitem[\protect\citeauthoryear{Van~Craenendonck and Blockeel}{2017}]{craenendonck2017constraint}
\begin{barticle}
\bauthor{\bsnm{Van~Craenendonck}, \binits{T.}},
\bauthor{\bsnm{Blockeel}, \binits{H.}}:
\batitle{Constraint-based clustering selection}.
\bjtitle{Machine Learning}
\bvolume{106}(\bissue{9}),
\bfpage{1497}--\blpage{1521}
(\byear{2017})
\doiurl{10.1007/s10994-017-5643-7}
\end{barticle}
\endbibitem

\bibitem[\protect\citeauthoryear{von Luxburg et~al.}{2012}]{luxburg2012clustering}
\begin{bchapter}
\bauthor{\bsnm{Luxburg}, \binits{U.}},
\bauthor{\bsnm{Williamson}, \binits{R.C.}},
\bauthor{\bsnm{Guyon}, \binits{I.}}:
\bctitle{Clustering: Science or art?}
In: \beditor{\bsnm{Guyon}, \binits{I.}},
\beditor{\bsnm{Dror}, \binits{G.}},
\beditor{\bsnm{Lemaire}, \binits{V.}},
\beditor{\bsnm{Taylor}, \binits{G.}},
\beditor{\bsnm{Silver}, \binits{D.}} (eds.)
\bbtitle{Proceedings of ICML Workshop on Unsupervised and Transfer Learning}.
\bsertitle{Proceedings of Machine Learning Research},
vol. \bseriesno{27},
pp. \bfpage{65}--\blpage{79}.
\bpublisher{PMLR},
\blocation{Bellevue, Washington, USA}
(\byear{2012}).
\burl{https://proceedings.mlr.press/v27/luxburg12a.html}
\end{bchapter}
\endbibitem

\bibitem[\protect\citeauthoryear{Van~Mechelen et~al.}{2023}]{mechelen2023white}
\begin{barticle}
\bauthor{\bsnm{Van~Mechelen}, \binits{I.}},
\bauthor{\bsnm{Boulesteix}, \binits{A.-L.}},
\bauthor{\bsnm{Dangl}, \binits{R.}},
\bauthor{\bsnm{Dean}, \binits{N.}},
\bauthor{\bsnm{Hennig}, \binits{C.}},
\bauthor{\bsnm{Leisch}, \binits{F.}},
\bauthor{\bsnm{Steinley}, \binits{D.}},
\bauthor{\bsnm{Warrens}, \binits{M.J.}}:
\batitle{A white paper on good research practices in benchmarking: The case of cluster analysis}.
\bjtitle{WIREs Data Mining and Knowledge Discovery}
\bvolume{13}(\bissue{6}),
\bfpage{1511}
(\byear{2023})
\doiurl{10.1002/widm.1511}
{\href{https://arxiv.org/abs/https://wires.onlinelibrary.wiley.com/doi/pdf/10.1002/widm.1511}{{https://wires.onlinelibrary.wiley.com/doi/pdf/10.1002/widm.1511}}}
\end{barticle}
\endbibitem

\bibitem[\protect\citeauthoryear{Yao et~al.}{2019}]{yao2019taking}
\begin{botherref}
\oauthor{\bsnm{Yao}, \binits{Q.}},
\oauthor{\bsnm{Wang}, \binits{M.}},
\oauthor{\bsnm{Chen}, \binits{Y.}},
\oauthor{\bsnm{Dai}, \binits{W.}},
\oauthor{\bsnm{Li}, \binits{Y.-F.}},
\oauthor{\bsnm{Tu}, \binits{W.-W.}},
\oauthor{\bsnm{Yang}, \binits{Q.}},
\oauthor{\bsnm{Yu}, \binits{Y.}}:
Taking human out of learning applications: A survey on automated machine learning
(2019)
{\href{https://arxiv.org/abs/1810.13306}{{arXiv:1810.13306}}}
{[cs.AI]}
\end{botherref}
\endbibitem

\bibitem[\protect\citeauthoryear{Zellinger and Bühlmann}{2023}]{zellinger2023repliclust}
\begin{botherref}
\oauthor{\bsnm{Zellinger}, \binits{M.J.}},
\oauthor{\bsnm{Bühlmann}, \binits{P.}}:
repliclust: Synthetic Data for Cluster Analysis
(2023)
\end{botherref}
\endbibitem

\bibitem[\protect\citeauthoryear{Z\"{o}ller and Huber}{2021}]{zoller2021benchmark}
\begin{barticle}
\bauthor{\bsnm{Z\"{o}ller}, \binits{M.-A.}},
\bauthor{\bsnm{Huber}, \binits{M.F.}}:
\batitle{Benchmark and survey of automated machine learning frameworks}.
\bjtitle{J. Artif. Int. Res.}
\bvolume{70},
\bfpage{409}--\blpage{472}
(\byear{2021})
\doiurl{10.1613/jair.1.11854}
\end{barticle}
\endbibitem

\end{thebibliography}

\end{document}